\documentclass[runningheads]{llncs}
\usepackage{import}
\usepackage{packages}

\begin{document}

\title{Leveraging Conditional Generative Models in a General Explanation Framework of Classifier Decisions}
\titlerunning{Conditional Generative Models in a General Explanation Framework}
\author{Martin Charachon\inst{1, 2} \and
Paul-Henry Cournède\inst{2} \and
Céline Hudelot\inst{2} \and 
Roberto Ardon\inst{1}}

\authorrunning{M. Charachon et al.}
\institute{Incepto Medical, France \and
MICS, Université Paris-Saclay, CentraleSupélec, France}

\maketitle  
\begin{abstract}
Providing a human-understandable explanation of classifiers' decisions has become imperative to generate trust in their use for day-to-day tasks.
Although many works have addressed this problem by generating visual explanation maps, they often provide noisy and inaccurate results forcing the use of heuristic regularization unrelated to the classifier in question. 
In this paper, we propose a new general perspective of the visual explanation problem overcoming these limitations. We show that visual explanation can be produced as the difference between two generated images obtained via two specific conditional generative models. Both generative models are
trained using the classifier to explain and a database to enforce the following properties: 
(i) All images generated by the first generator are classified similarly to the input image, whereas the second generator's outputs are classified oppositely.
(ii) Generated images belong to the distribution of real images.
% (ii) For each of the two generative models, the Wasserstein-2 distance between the distributions of generated and real images is minimal.
(iii) The distances between the input image and the corresponding generated images are minimal so that the difference between the generated elements only reveals relevant information for the studied classifier. Using symmetrical and cyclic constraints, we present two different approximations and implementations of the general formulation. Experimentally, we demonstrate significant improvements w.r.t the state-of-the-art on three different public data sets. In particular, the localization of regions influencing the classifier is consistent with human annotations. 

\end{abstract}
% \documentclass[../main.tex]{subfiles}
% \graphicspath{{\subfix{images/}}}

% \begin{document}

%----------------------------------------------------------------------------------------
%	Introduction
%----------------------------------------------------------------------------------------

% \begin{figure}[ht]
% \begin{center}
% \centerline{\includegraphics[width=0.8\textwidth]{images/sahm_xr_br_d_v5.png}}
% \caption{\textbf{Adversary, stable and visual explanation generations}. From left to right for the three considered classification problems (i.e. MNIST digit classification, Chest X-Rays pneumonia detection and Brain MRI tumor localization): the original image; the ground annotations pointing out the relevant regions for humans; the adversarial image (for the classifier to explain); the stable image; our resulting explanation map of the classifier's decision.}
% \label{fig:sahm_xr_br_d_v5}
% \end{center}
% \end{figure}

\section{Introduction}\label{sec:introduction}
Deep learning (DL) models represent the state-of-the-art for many computer vision tasks, in particular in the medical image domain \cite{mrnet,skincancerEsteva2017}.
Although increasingly accurate, DL classifications models are still missing a general framework providing human readable explanation of their results. 
In radiology (as for other critical fields), human conclusions on images are generally communicated to peers with explanations. It aims to increase confidence by stimulating criticism or approval. Similarly, it is imperative to provide explanations for DL classification results. Their adoption in sensitive fields such as clinical practice is at stake.
Many contributions have been made on the topic of DL interpretability and explainability (see sec. \ref{sec:related_work}).
Numerous methods, based on back-propagation techniques, 
%\cite{simonyan_deep_2014,springenberg_striving_2014, smilkov_smoothgrad:_2017, sundararajan_axiomatic_2017}
last network layers analysis
%\cite{Zhou2016LearningDF,selvaraju_grad-cam:_2017,rajpurkar_chexnet:_2017}
or input perturbations,
%\cite{fong_interpretable_2017, dhurandhar_explanations_2018, hsieh2020evaluations, Fong2019UnderstandingDN}
produce visual explanation maps through a process that only depends on the single input image to the classifier. 
%for which explanation on the model's output is sought. 
These methods are often not model-agnostic and/or need regularization heuristics to produce visualization maps acceptable to humans.
They require significant manual adaptations when changing the DL model, especially if the application domain changes (e.g. from natural to medical images).

Our method produces a visual explanation as the difference between a stable and an adversarial generation. 
This idea is also used in \cite{Charachon2020CombiningSA}, but we adopt a different point of view inspired by counterfactual perturbations \cite{chang_explaining_2019} and domain translation \cite{Narayanaswamy2020ScientificDB,CycleGAN2017}: in our approach both generators are trained to produce images that are within the distribution of real input images.
% Inspired by counterfactual perturbations \cite{chang_explaining_2019} and domain translation \cite{Narayanaswamy2020ScientificDB,CycleGAN2017}, our method builds upon \cite{Charachon2020CombiningSA} improving its formulation and results.
% Inspired by \cite{chang_explaining_2019,Narayanaswamy2020ScientificDB,CycleGAN2017}, our method builds upon \cite{Charachon2020CombiningSA} improving its formulation and results. 
% We consider the same approach in the definition of the visual explanation but both generators are now trained to produce images that are within the distribution of real input images.
% In particular, the adversary is searched as the closest element to the input image within the distribution of real images classified differently. While close in spirit, this formulation implies a completely different implementation allowing the adversary to capture crucial patterns for the classifier \textbf{within} the distribution of real images. 
In particular, the adversary is searched as the closest element to the input image within the distribution of real images classified differently.
% Thus, it captures crucial patterns for the classifier.
Our contributions are as follows: \textbf{(i)} A new formal definition of the visual explanation task as a constrained optimization problem (sec. \ref{sec:general_formulation}). \textbf{(ii)} Two approximations and implementations using cyclic constraints, one being simpler than the other but with slightly inferior results (sec. \ref{sec:SyCE} and \ref{sec:CyCE}). \textbf{(iii)} A proposal on how to compare the capacity of two generators in producing images within a target distribution (sec. \ref{sec:metrics_DT}).

We validate our method on three publicly available data sets: numbers classification on MNIST, detection of pathological chest X-Rays, and brain tumors in MRI (sec. \ref{sec:experiments}). Our results outperform state-of-the-art methods (sec. \ref{sec:results}).

% \end{document}
% \documentclass[../main.tex]{subfiles}
% \graphicspath{{\subfix{images/}}}

% \begin{document}

%----------------------------------------------------------------------------------------
%	Related Work
%----------------------------------------------------------------------------------------
\section{Related Works}\label{sec:related_work}

% \subsection{Visual Explanation Methods}\label{subsec:VisualExplanationMethods}
\textbf{Visual Explanation Methods }Early works propose to analyze either backpropagation of gradients  \cite{simonyan_deep_2014,springenberg_striving_2014,smilkov_smoothgrad:_2017,sundararajan_axiomatic_2017} or last layers activation \cite{Zhou2016LearningDF,selvaraju_grad-cam:_2017} w.r.t a given input.
While providing reasonable outputs in some settings, these methods often present a series of drawbacks, ranging  from producing noisy explanation maps \cite{simonyan_deep_2014} to not being model-agnostic \cite{Zhou2016LearningDF}. In the worst scenarios, they have been found independent of both model parameters and label randomization \cite{adebayo_sanity_2018}.

\textit{Perturbation methods- }In contrast, perturbation-based methods study the impact on the model's output of perturbations applied on the input image. They typically compute an optimal binary mask ($M$) that determines where a perturbation function ($\Phi$) should act on the input image to change the classifiers' output. The visual explanation map is then given by $M$. 
For each image, \cite{fong_interpretable_2017} introduces an optimization setting to find the minimal perturbation region with the greatest impact on the classifier's output.
Building on this work, \cite{dhurandhar_explanations_2018,hsieh2020evaluations,Fong2019UnderstandingDN} propose slightly different image-wise optimization problems, whereas \cite{dabkowski_real_2017} train a neural network to generate such perturbation region on a given database.
To produce acceptable visual results, these methods impose strong and specific regularization that are unrelated to the classifier to interpret.

\textit{Adversarial Examples as Explanation-} Rather than using a perturbation function and optimizing a mask, \cite{woods_adversarial_2019,Elliott2019AdversarialPO} propose to find for each input image an $L_2$-close adversarial example (to be compared to the input) that impacts the classifier's decision within a constrained space. More recently, \cite{Charachon2020CombiningSA} train two models with the same architecture and partly shared parameters to produce for a given input image, $L_1$,$L_2$-close images with respectively similar and opposite classifications compared to the original one.
The visual explanation is then defined as the difference between the two generated images.
These approaches avoid mask-related regularization heuristics. Our general formulation shares some similarities with the work of \cite{Charachon2020CombiningSA}, especially in the use of a stable generation to improve regularity. Yet there are clear conceptual differences. First, in their formulation, nothing constrain generations (adversarial in particular) to be consistent with real distributions. Second, they impose an element-wise distance constraint ($L_2, L_1$ distances) between the adversary and the input image. This overconstrains the adversarial generation and tends to produce adversarial artefacts. It also prevents the capture of distribution specific patterns. We will show that our formulation solves these issues.

\textit{Counterfactual Explanation- }Built upon \cite{fong_interpretable_2017} or \cite{dabkowski_real_2017}, \cite{chang_explaining_2019,major_interpreting_2020,Lenis2020DomainAM} propose to generate realistic perturbations with generative models attached to the input domain, e.g. perturbing by healthy tissue an image classified as pathological. However, strong regularizations on the perturbation regions are still needed. Instead, \cite{Goyal2019CounterfactualVE,Wang2020SCOUTSD} use the minimal difference between the input image and a counterfactual images in the database. Although they capture relevant information for the classifier within the distribution of real images, they derive heuristically a region of interest and they still require a counterfactual image (with a completely different structure) to which they can compare the input image.\\

% \subsection{Domain Translation with GANs}\label{subsec:DomainTranslation}
\noindent\textbf{Domain Translation with GANs }Independently, image-to-image translation is a common and successful application of Generative Adversarial Networks (GANs) \cite{NIPS2014_5423}. It consists in learning a mapping between two different image domains, whether we have access to paired data \cite{pix2pix2017} or not \cite{CycleGAN2017,Liu2017UnsupervisedIT,shen_one--one_2020}.
In particular, CycleGAN \cite{CycleGAN2017} designed for unpaired image-to-image translation, introduces a cycle-consistency constraint that enforces a certain proximity  across domains. In medical imaging, a CycleGAN framework is used in \cite{Narayanaswamy2020ScientificDB} to emphasize important structures that differ between images from different classes. However, this framework does not interpret a classifier's decision, but rather gives additional insights on what is different between healthy and pathological images. 
Another line of works directly builds interpretable classifiers \cite{Bass2020ICAMIC,seah_chest_2019} or generations \cite{Baumgartner2018VisualFA,Wolleb2020DeScarGANDA} using this idea of domain translation. 
Finally, through latent space conditioning in \cite{Singla2020ExplanationBP}, the authors propose a framework to align progressive plausible generations with changes of classifier's prediction score. Note that with some adjustments, we could use their framework as an embodiment of our general formulation.\\

\section{General Formulation}\label{sec:general_formulation}
We formulate our method in the case of a binary classification problem (see \textbf{suppl. section 2} for multi-class adaptation). 
Let $\reals$ denote the space of real images. We define the visual explanation $\Expl$ of a classifier $\f$ on $x\in\reals$ as the difference between two generated images $\gs(x)$ and $\ga(x)$. $\gs(x)$, the stable generation, is built to be classified as $x$. $\ga(x)$, the adversary generation, is built to be classified oppositely.
% Let $\cf(x) \in \{0,1\}$ be the class of $x$ predicted by $\f$. We have:
The visual explanation $\Expl$ thus reads:
\begin{equation}\label{eq:ExplanationDef}
	\Expl(x) = |\gs(x) - \ga(x)|
\end{equation}
% where $\gs$ and $\ga$ are the stable and adversary generators respectively. 
We claim that, to be acceptable, $\Expl$ should only capture relevant, regular and consistent information on $x$ impacting the decision of $\f$. This translates into the following conditions on $\gs$ and $\ga$:\\
\textbf{Relevance- }$\gs(x)$ and $\ga(x)$ should only differ in regions that are relevant to the classification of $x$ by $\f$. \\
 \textbf{Regularity- }
Both generation process should be comparable in order to avoid differences independent from the classifier $\f$ (residual noise imputable to generation process).
%  $x$ and $\ga(x)$ are not directly comparable as there exists a residual noise imputable to operations performed on $x$ through $\ga$. We assume that if the generation processes are comparable, then this noise will tend to be eliminated. The authors of \cite{Charachon2020CombiningSA} experimentally verified this hypothesis.\
\\
%  \textbf{Regularity- }$\gs$ and $\ga$ should belong to a set of functions ($\Fam$) so that $\Expl$, defined as their difference, is perceived as smooth. This property is important for human acceptation of the generated explanation map.\\
\textbf{Consistency with reality- }$\gs(x)$ and $\ga(x)$ should belong to the distribution of real images.
This last property is essential to avoid adversarial generated artifacts (typical of adversarial attacks). It also reveals distribution-specific patterns influencing $\f$ only visible if attacks are coherent with the distribution of real images \footnote{Fig. \ref{fig:dt_xr_br} shows that, unlike other methods, our adversarial generations produce only real specific differences between a "3" and a "8" (and vice-versa).}.

Summarizing these conditions, $\gs$ and $\ga$ are searched as a solution couple of the following optimization problem:
\begin{equation}\label{eq:paperformulation}	
	(\gs, \ga) = \hspace{-1.6cm}
	\underset{
			\begin{array}{c}
				\scriptstyle \gsopt, \gaopt \\
				\scriptstyle \\
			    \underbrace{
				\scriptstyle \text{s.t. } 
				\left\{		
				% \hspace{-1mm}
				\begin{array}{c}
					\scriptstyle \gsopt(\chi_0) \subset \chi_0, \gsopt(\chi_1) \subset \chi_1 \\
					\scriptstyle \gaopt(\chi_0) \subset \chi_1, \gaopt(\chi_1) \subset \chi_0
				\end{array}
				% \hspace{-1mm}
				\right\} }_{\scriptstyle \text{\color{blue} \textbf{Consistency with reality}}}
			\end{array} }
		{\mathrm{argmin}} \hspace{-1.3cm}	
		\overbrace{
		\Big[
		\small
		\underbrace{d_g(\gsopt, \gaopt)}_{
		\text{\color{blue} \textbf{Regularity} }} +
		\mathbb{E}_x   \left (d_s(x, \gsopt(x)) + d_a(x, \gaopt(x)) 
		\right )\Big]
		}^{\text{\color{blue} \textbf{Relevance}}} 	
\end{equation}
(i) $\chi_0$ and $\chi_1$ are the subsets of real images classified as class \textbf{0} and \textbf{1} respectively by $\f$, they form a partition of $\reals$;\\
(ii) $(d_a, d_s)$ are distances measuring the proximity to $x$ of each generated image;\\
(iii) $d_g$ is a distance between generators, its minimization aims to eliminate errors inherited from generation processes and irrelevant to $\f$.
% (iii) $d_g$ is a distance between generators, its minimization aims to eliminate generation errors irrelevant to $\f$.\\

In the next sections, the implementations of two different approximations of this formulation are detailed.
\section{SyCE: Symmetrically Conditioned Explanation} \label{sec:SyCE}
% We propose a concrete and practical embodiment of problem \eqref{eq:paperformulation} that leverages ideas from \cite{CycleGAN2017,shen_one--one_2020} and alleviates some limitations of \cite{Charachon2020CombiningSA}. 
\subsection{Generator definition using Symmetry}\label{subsec:SyCE_formulation}
\newcommand\gz{{g_0^{\star}}}
\newcommand\gu{{g_1^{\star}}}
% Among the terms in \eqref{eq:paperformulation}, $d_g$ and $d_a$ are challenging to concretely grasp and approximate.

% In the following embodiment, instead of introducing a constraint on generators proximity ($d_g$ in \eqref{eq:paperformulation}), we propose a built-in regularization.
In the following embodiment we propose a built-in proximity between generators. We avoid the difficulty of explicitly introducing a proximity term $d_g$ in \eqref{eq:paperformulation}  that depends of the choice of generators \footnote{In \cite{Charachon2020CombiningSA}, the $L_2$ distance between network parameters is penalized but may not be adapted to another type of generators.}.
The idea is to force both adversarial and stable generations to lay in the image space of a unique generator. $\ga$ being the adversarial generator, this can be achieved by imposing on the stable generator $\gs = \ga\circ \ga$. 
Both generated images are then the result of the same generation process and their difference is less subject to purely reconstruction errors.

Moreover, since we expect $\gs(x)$ to be the closest element to $x$ produced by the generation process, $\gs(x) \approx x$ should be constrained. Combined with the built-in proximity between generators, this induces a symmetry constraint on $\ga$: $\ga^2(x) \approx x$. As supported by empirical results in
\cite{CycleGAN2017,shen_one--one_2020}, adversarial generation $\ga$ should also be in the proximity of $x$. The symmetry constraint compels $\ga$ to transpose elements from one classification space to the other ($\chi_0 \Longleftrightarrow \chi_1$) and "easily" return. The embodiment of problem \eqref{eq:paperformulation} then reads

\begin{equation}\label{eq:SSyE}
	\ga  = \hspace{-0.9cm}
	\underset{
		\begin{array}{c}
			\scriptstyle g_{f_c}^a \  \  \text{s.t.} \\
			\scriptstyle g_{f_c}^a(\chi_0) \subset \chi_1, g_{f_c}^a(\chi_1) \subset \chi_0\\
			\scriptstyle g_{f_c}^{a2}(\chi_0) \subset \chi_0, g_{f_c}^{a2}(\chi_1) \subset \chi_1
		\end{array}
	}{\mathrm{argmin}} \hspace{-0.9cm}
	\begin{array}{c}
		\mathbb{E}_{x \in \chi} \left (||x -  g_{f_c}^{a2}(x)||_{1,2} \right ) 
	\end{array}
\end{equation}
 where we implicitly minimize $d_g$ and $d_a$ through the optimization constraints and explicitly set and optimize $d_s$ using a combination of $L_1$ and $L_2$ (see \textbf{suppl. section 1}).

In practice, to train a unique generator results in a too restrictive setting. Adversarial and stable generations are in general too close and the adversary fails to fall into the opposite  distribution (see \textbf{suppl. section 6.1}). 

% Such constrained generation process implicitly enforces the minimization of $d_a$. 

% In contrast with \cite{Charachon2020CombiningSA}, this constraint does not penalize $\gz(\chi_0) \subset \chi_1$ and $\gu(\chi_1) \subset \chi_0$.

% Since we seek to keep both generation "close" to the input image ($(d_a,d_s)$), the use of cycle consistancy between generators as in \cite{CycleGAN2017, shen_one--one_2020}  If $\ga(x)$ defines the adversarial generation, we propose to constrain the stable generator to be $g^2(x)$ (twice applying $g$ on the input image).

% The following choices aim to do so.
To alleviate this, we relax the formulation and search for two auxiliary generators $\gz$ and $\gu$ defined on $\reals$, but verifying the above properties only on elements of $\chi_0$, respectively $\chi_1$. Thus, they should satisfy symmetry constraints respectively on $\chi_0$ and $\chi_1$:
\begin{equation}\label{eq:symmetry}
\left[ \forall x \in \chi_0,\  {\gz}^2(x) \approx x  \right]; 
\left[ \forall x \in \chi_1,\ {\gu}^2(x) \approx x  \right]
\end{equation}

\begin{figure}[ht]
    \begin{center}
    \centerline{\includegraphics[width=0.7\textwidth]{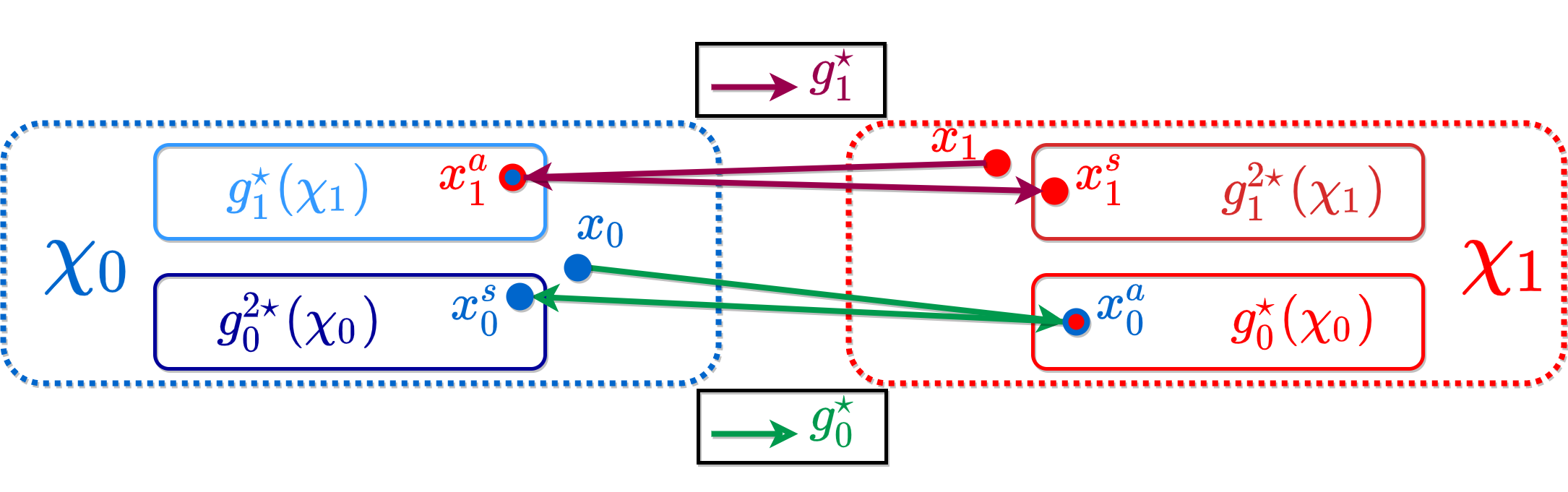}}
    \end{center}
\caption{\textbf{Generators mappings in SyCE.} $\gz$ maps each element $x_0 \in \chi_0$ to an element $x_0^a \in \gz(\chi_0) \subset \chi_1$. By reapplying $\gz$, given the symmetry constraint, this element  $x_0^a \in \gz(\chi_0)$ is mapped back to an element $x_0^s \in \gz^2(\chi_0) \subset \chi_0$ very close to the original image $x_0$. $\gu$ acts similarly on $x_1 \in \chi_1$.}
\label{fig:syce_schema}
\end{figure}

Figure \ref{fig:syce_schema} gives an illustration of the mappings built by $(\gz, \gu)$. We can then define adversarial and stable generations as 
\begin{equation}\label{eq:sym_adversary}
    \begin{array}{cc}
    	\ga(x) = \left \{
    	\begin{array}{l}
    		\gz(x) \in \chi_1 \text{ if } x\in\chi_0 \\
    	    \gu(x) \in \chi_0 \text{ if } x\in\chi_1 \\
    	\end{array}\hspace{-0.1cm} 
    	\right. ;
    	\   \  &
    	\gs(x) = \left \{
    	\begin{array}{l}
    		\gz^2(x) \in \chi_0 \text{ if } x\in\chi_0\\
    		\gu^2(x) \in \chi_1 \text{ if } x\in\chi_1\\
    	\end{array}\hspace{-0.1cm} 
    	\right.
    \end{array}
\end{equation}

And the relaxed version of \eqref{eq:SSyE} reads
\begin{equation}\label{eq:SyCE}
(\gz, \gu ) = \hspace{-0.9cm}
	\underset{
		\begin{array}{c}
			\scriptstyle g_0, g_1 \\
			\scriptstyle \text{s.t.} \\
			\scriptstyle g_0(\chi_0) \subset \chi_1, g_1(\chi_1) \subset \chi_0\\
			\scriptstyle g_0^2(\chi_0) \subset \chi_0, g_1^2(\chi_1) \subset \chi_1
		\end{array}
	}{\mathrm{argmin}} \hspace{-0.6cm}
	\left[  
	\begin{array}{c}
		\mathbb{E}_{x \in \chi_0} \left (||x - g_0^2(x)|| \right )  \\  + \\
		\mathbb{E}_{x \in \chi_1} \left (||x - g_1^2(x)||\right )
	\end{array}
	 \right]
\end{equation}

\noindent Finally, the expression of visual explanation is 
\begin{equation}\label{eq:ExplanationDefg0g1}
	\Expl(x) = 
	\left \{\begin{array}{cc}
		|\gz^2(x) - \gz(x)|\, \text{ if }\,  x \in \chi_0 \\
		|\gu^2(x) - \gu(x)|\, \text{ if }\,  x \in \chi_1
	\end{array}
	\right \}
\end{equation}

% \begin{figure}[ht]
%     \begin{subfigure}[h]{0.36\textwidth}
%     \caption{Generators mappings}
%     \begin{center}
%     \centerline{\includegraphics[width=\textwidth]{images/generators_mappings_ecml_v2.png}}
%     \label{fig:SyCE_shcema}
%     \end{center}
%     \vskip -0.5cm
%     \end{subfigure}
%     \begin{subfigure}[h]{0.63\textwidth}
%     \caption{Optimization framework for $x \in \chi_1$}
% 	\begin{center}
% 		\centerline{\includegraphics[width=\textwidth]{images/syce_ecml_10.png}}
% % 		\caption{\textbf{Overview of the SyCE optimization framework}. Training step of $g_1$ and $g_0$ for an original image $x \in \chi_1$. The terms $L_i^1$ are the $L_i$ parts that  act on $x \in \chi_1$. \textbf{\textit{Note}}: the training step is similar for an original image $x \in \chi_0$ with corresponding loss terms $L_i^0$.}
% 		\label{fig:syce_ecml_10}
% 	\end{center}
% 	\end{subfigure}
% \vskip -0.2cm
% \caption{\textbf{Overview of SyCE.} \textbf{\textit{(a) Generators mappings}}: $\gz$ maps each element $x_0 \in \chi_0$ to an element $\gz(x_0) \in \gz(\chi_0) \subset \chi_1$. By reapplying $\gz$, given the symmetry constraint, this element  $\gz(x_0) \in \gz(\chi_0)$ is mapped back to an element in $\gz^2(\chi_0) \subset \chi_0$ very close to the original image $x_0$. $\gu$ acts similarly on $x_1 \in \chi_1$. \textbf{\textit{(b) Training step}} of $g_1$ and $g_0$ for an original image $x_1 \in \chi_1$. The terms $L_i^1$ are the $L_i$ parts that act on $x_1 \in \chi_1$. Similar training step for $x_0 \in \chi_0$.}
% \end{figure}
\subsection{Weak Formulation}\label{subsec:FrameworkOverview}

Formulation \eqref{eq:SyCE} can be further approximated into an unconstrained min-max optimization problem commonly used to optimize GANs \cite{NIPS2014_5423}.
\begin{equation}\label{eq:minmaxpb}
	\underset{g_0, g_1}{\mathrm{min}}\underset{D_0, D_1}{\mathrm{max}}\hspace{0.1cm} L_{Tot}(x, g_0, g_1, D_0, D_1)
\end{equation}
where  $D_0$ (resp. $D_1$) is a domain specific discriminator in charge of distinguishing real images in $\chi_0$ (resp. $\chi_1$) from outputs of $g_1$ (resp. $g_0$). $L_{Tot}$ is the loss function necessary to weakly approximate \eqref{eq:SyCE} and whose components are now detailed.
We denote {\small $\cf(x) \in \{0,1\}$} the class of $x$ predicted by $\f$ and obtained from $\f(x)$ by a threshold.

\begin{figure}[ht]
	\begin{center}
	\centerline{\includegraphics[width=0.9\textwidth]{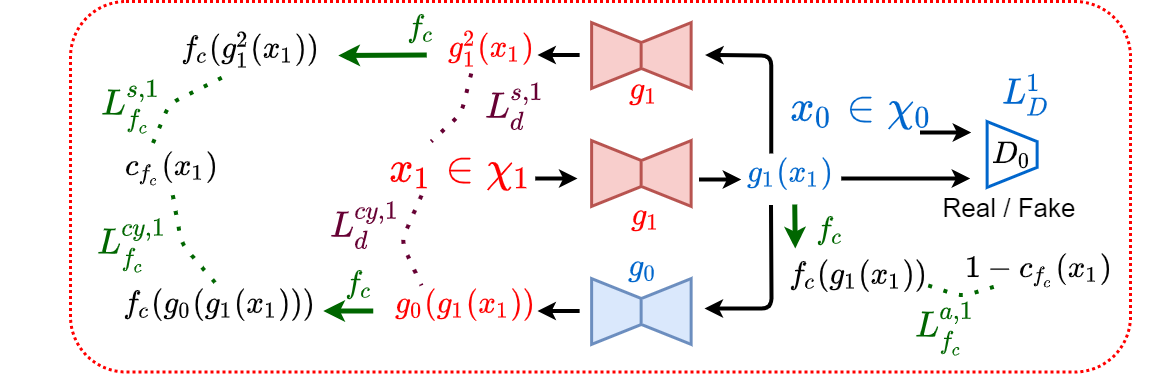}}
	\end{center}
\caption{\textbf{Overview of the SyCE optimization framework}. Training step of $g_1$ and $g_0$ for an original image $x_1 \in \chi_1$. The terms $L_i^1$ are the $L_i$ parts that  act on $x_1 \in \chi_1$. \textbf{\textit{Note}}: the training step is similar for an original image $x_0 \in \chi_0$ with corresponding loss terms $L_i^0$.}
\label{fig:syce_ecml_10}
\end{figure}

\noindent\textbf{Adversarial Generation-}
For a real image $x\in\chi_0$, $g_0(x)$ should be classified as of class \textbf{1}, and reciprocally, if $x$ is a real image in $\chi_1$, $g_1(x)$ should be classified as of class \textbf{0} by $\f$. We thus introduce into $L_{Tot}$ the term:
\begin{equation}\label{eq:loss_fcadv}
	\begin{array}{lclc}
		 L_{f_c}^{a}(x, g_0, g_1) &=&  \mathbb{E}_{x \in \chi_0} L_{bce}(1 - c_{f_c}(x), f_c(g_0(x))) & + \\
		&& \mathbb{E}_{x \in \chi_1} L_{bce}(1 - c_{f_c}(x), f_c(g_1(x))) 
	\end{array}
\end{equation}
where $L_{bce}$ is the binary cross entropy loss function. This term constitutes a typical "attack" on classifier $\f$ and does not enforce generated images to belong to the distributions of "real" images in $\chi_0$ or $\chi_1$. To cope with this, we introduce into $L_{Tot}$ a classical GAN term
\begin{equation}\label{eq:loss_D}
	\begin{array}{lclc}
		L_{D}(x, g_0, g_1, D_0, D_1) &=& \mathbb{E}_{x \in \chi_0} \hspace{0.1cm} [ \hspace{0.025cm} L_{bce}(1, D_1(x)) + L_{bce}(0, D_1(g_0(x))) ] &  + \\
		&& \mathbb{E}_{x \in \chi_1} \hspace{0.1cm} [ \hspace{0.025cm} L_{bce}(1, D_0(x)) + L_{bce}(0, D_0(g_1(x))) ] 
	\end{array}
\end{equation}
where $D_0$ and $D_1$ are trained to minimize it while $g_0$ and $g_1$ to maximize it. 

\noindent\textbf{Symmetry-}
Symmetry objectives of problem \eqref{eq:SyCE} are directly enforced in $L_{Tot}$ by training $g_0$ and $g_1$ to minimize 
\begin{equation}\label{eq:loss_dst}
L_{d}^{s}(x, g_0, g_1) = \mathbb{E}_{x \in \chi_0} \|x - g_0^2(x) \|_{1,2} + \mathbb{E}_{x \in \chi_1} \|x - g_1^2(x)\|_{1,2} 
\end{equation}
Using the average of $L_1$ and $L_2$ distances produces better results in our experiments. Additionally, to drive the classification of generated elements $g_0^2(x)$ and $g_1^2(x)$ towards $\cf(x)$, we add the loss term:  
\begin{equation}\label{eq:loss_fcst}
	\begin{array}{lclc}
		L_{f_c}^{s}(x, g_0, g_1) &=&  
		 \mathbb{E}_{x \in \chi_0} L_{bce}(c_{f_c}(x), f_c(g_0^2(x))) & + \\
		&& \mathbb{E}_{x \in \chi_1} L_{bce}(c_{f_c}(x), f_c(g_1^2(x))) 
	\end{array}
\end{equation}

\noindent\textbf{Cyclic Consistency- }Constraints in \eqref{eq:SyCE} only induce $g_0 \circ g_1(\chi_1) \subset \chi_1$ and $g_1 \circ g_0(\chi_0) \subset \chi_0$. In practice, enforcing a stronger relation between $g_0$ and $g_1$ increases convergence speed and encourages the minimisation of $d_g$ in \eqref{eq:paperformulation}. As in \cite{CycleGAN2017}, we introduce in $L_{Tot}$
\begin{equation}\label{eq:loss_dcyc}
L_{d}^{cy}(x, g_0, g_1) = \mathbb{E}_{x \in \chi_0} \|x - g_1(g_0(x)) \|_{1,2}  + \mathbb{E}_{x \in \chi_1} \|x - g_0(g_1(x)) \|_{1,2} 
\end{equation}
and thus couple optimizations of $g_0$ and $g_1$. Finally, to also ensure that cyclic terms are classified as $x$ we add consistency loss
\begin{equation}\label{eq:loss_fccyc}
	\begin{array}{lclc}
		L_{f_c}^{cy}(x, g_0, g_1) &= &
		 \mathbb{E}_{x \in \chi_0} L_{bce}(c_{f_c}(x), f_c(g_1(g_0(x))) &  + \\
		&&\mathbb{E}_{x \in \chi_1} L_{bce}(c_{f_c}(x), f_c(g_0(g_1(x))) 
	\end{array}
\end{equation}
In experiments, we observe that $L_{f_c}^{cy}$ has a smaller effect than $L_{f_c}^{s}$ yet it slightly improves the cycle consistency.

The global loss function of the optimization problem \eqref{eq:minmaxpb} then reads
\begin{equation}\label{eq:total_loss}
	L_{Tot} = \lambda_{d}^{s} L_{d}^{s} + \lambda_{d}^{cy} L_{d}^{cy} + \lambda_{f_c}^{a} L_{f_c}^{a}  + \lambda_{f_c}^{s} L_{f_c}^{s} +  \lambda_{f_c}^{cy} L_{f_c}^{cy} + \lambda_{D} L_{D}
\end{equation}
% \begin{equation}\label{eq:total_loss}
% 	L_{Tot} = 
% 	\left\{ 
% 	\begin{array}{ll}
% 		\hspace{-0.1cm} \lambda_{d}^{s} L_{d}^{s} + \lambda_{d}^{cy} L_{d}^{cy} + &\\
% 		\hspace{-0.1cm} \lambda_{f_c}^{a} L_{f_c}^{a}  + \lambda_{f_c}^{s} L_{f_c}^{s} +  \lambda_{f_c}^{cy} L_{f_c}^{cy} + &\\
% 		\hspace{-0.1cm} \lambda_{D} L_{D}&
% 	\end{array}
% 	\right\}
% \end{equation}
where parameters $ \lambda_{d}^{s}, \lambda_{d}^{cy}, \lambda_{f_c}^{a}, \lambda_{f_c}^{s}, \lambda_{f_c}^{cy}$ and $\lambda_{D} \in \mathbb{R}^{+}$ control the relative importance of the different terms. Figure \ref{fig:syce_ecml_10} shows an overview of the whole framework.

\section{CyCE: Cyclic Conditioned Explanation}\label{sec:CyCE}
Inspired by the CycleGAN framework used in \cite{Narayanaswamy2020ScientificDB,Wolleb2020DeScarGANDA}, we also propose a simpler embodiment of the general formulation \eqref{eq:paperformulation}. As in equation \eqref{eq:sym_adversary}, we introduce generative models $g_{0}^{\star}$ and $g_{1}^{\star}$ to define the conditional adversary generator $\ga$. 
In contrast, we relax the formulation and define $\gs$ as the identity. The visual explanation becomes ${\cal E}(x) = |x - g_{0}^{\star}(x)|$ if $x \in \chi_0$, ${\cal E}(x) = |x - g_{1}^{\star}(x)|$ otherwise. In this formulation, we directly compare the adversary to the original image removing both constraints $d_g$ and $d_s$. We trade potential reconstruction errors against a better proximity of the generated elements to the adversary class.
Concerning $d_a$, the cycle consistency term encourages the proximity to the input. The approximated problem reads
\begin{equation}\label{eq:CyCE}
	(\gz, \gu ) = \hspace{-0.9cm}
	\underset{
		\begin{array}{c}
			\scriptstyle g_0, g_1\\
			\scriptstyle \text{s.t.} \\
			\scriptstyle g_0(\chi_0) \subset \chi_1, g_1(\chi_1) \subset \chi_0
		\end{array}
	}{\mathrm{argmin}} \hspace{-0.6cm}
	\left[  
	\small
	\begin{array}{c}
		\mathbb{E}_{x \in \chi_0} \left (||x -  g_1(g_0(x))|| \right )  \\  + \\
		\mathbb{E}_{x \in \chi_1} \left (||x -  g_0(g_1(x))||\right )
	\end{array}
	 \right]
\end{equation}
In practice, the optimization framework is very close to the one presented in Section \ref{subsec:FrameworkOverview}. We only remove the terms related to symmetrical elements $g_0^2$ and $g_1^2$ i.e. $L_d^s$ and $L_{f_c}^s$ (see \textbf{suppl. sections "Reminder" and 5.1}).

We have mainly presented two possible embodiments of problem \eqref{eq:paperformulation}, which will be compared in the following. Others are considered in the \textbf{suppl. section 1}. For instance, we further describe the single generator version introduced in section \ref{subsec:SyCE_formulation} (see equation \eqref{eq:SSyE}), and propose a cyclic variation of the work of \cite{Charachon2020CombiningSA}.

% \end{document}
% \documentclass[../main.tex]{subfiles}
% \graphicspath{{\subfix{images/}}}

% \begin{document}
%----------------------------------------------------------------------------------------
%	Experiments
%----------------------------------------------------------------------------------------

\section{Experiments}\label{sec:experiments}

\subsection{Datasets and Classifiers}\label{subsec:DatasetsandClassifiers}
\noindent\textbf{Digits Identification - MNIST- }
We designed a binary classification task on the MNIST datasets \cite{lecun-mnisthandwrittendigit-2010} that consists in distinguishing digits "3" from digits "8".
We extracted digits "3" and "8" from the original dataset to create training, validation and test sets of respectively 9585,  2397 and 1003 samples. The original images of size 28 x 28 are normalized to [0, 1]. We trained a convolutional network based on LeNet \cite{lenet} to minimize a binary cross entropy. The classifier reaches an AUC very close to 1.0 (and accuracy of 0.997) on the test set.

\noindent\textbf{Pneumonia detection - Chest X-Rays- }
We created a chest X-Rays dataset from the available RSNA Pneumonia Detection Challenge which consists of 26684 X-Ray dicom exams extracted from the NIH CXR14 dataset \cite{DBLP:journals/corr/WangPLLBS17}. As in \cite{Charachon2020CombiningSA}, we only kept the healthy and pathological exams constituting a binary database of 14863 samples (8851 healthy / 6012 pathological). Pathological cases are provided with bounding box annotations around opacities. We randomly split the dataset into train (80 \%), validation (10 \%) and test (10 \%) sets. 
A ResNet50 \cite{DBLP:journals/corr/HeZR016} and a DenseNet121 \cite{Huang2017DenselyCC} were trained to minimize a binary cross entropy loss, on images rescaled from 1024 x 1024 to 224 x 224 and normalized to [0, 1]. They respectively achieve 0.974 and 0.978 AUC scores on the test set. 

\noindent\textbf{Brain Tumor localization - MRI- }
Finally, we consider the problem of localizing brain tumor in Magnetic Resonance Imaging (MRI) volumes as a binary classification task. We propose to classify each slice (2D image) along the axial axis as containing either at least one tumor region (class "1"), or none (class "0"). The brain MRI dataset comes from the Medical Segmentation Decathlon Challenge \cite{Simpson2019ALA}. In this work, we only use the contrasted T1-weighted (T1gd) sequence and transform the expert multi-level annotations into binary mask annotation with which we obtain the class label of each slice. The train-validation-test split consists of 46900, 6184, and 9424 slice images of size 224 x 224. We also train a ResNet50 and a DenseNet121 with the same settings as the Chest X-Rays problem described above, except that they are trained to minimize a weighted binary cross entropy and achieve test set AUC values of 0.975 and 0.980.

For all the problems, we use the Adam optimizer \cite{DBLP:journals/corr/KingmaB14} with an initial learning rate of 1e-4. Random geometric transformations such as zoom, translations, flips or rotations are introduced during training. A more detailed presentation of the datasets, annotations and classifiers is given in the supplementary materials (see sections \textbf{3} and \textbf{4})

\subsection{Visual explainer implementation}\label{subsec:Implementation}
% \subsubsection{Our methods}\label{subsubsec:ours}
\noindent\textbf{Our methods- } For both SyCE and CyCE, the generators $g_0$ and $g_1$ have the same structure and follow a UNet-like \cite{DBLP:journals/corr/RonnebergerFB15} architecture as introduced in \cite{pix2pix2017} for image-to-image translation. The discriminators $D_0$ and $D_1$ consist of convolutional downsampling blocks followed by a dense linear layer. \\ 
The models are trained using Adam with an initial learning rate of 1e-4 for the generators and 2e-4 for the discriminators. In practice, we compute one optimization step for $g_0$ and $g_1$ given a batch of images $x$ in the source domain $\chi_0$, and target domain $\chi_1$. We proceed symmetrically after switching the source and the target domains for a batch of images in $\chi_1$. Then, we optimize the two discriminators. We elaborate more about the model architectures, the training implementation, and give the weighting parameters $\lambda_i$ in \textbf{supp. section 5.1}).
% as well as in the provided code of our experiments (available on github at \url{https://anonymous.4open.science/r/26e23a4f-14a0-43f6-9287-979847e2851a/}). 
 
\noindent\textbf{Comparison to Baselines- } We compare our proposed methods against several state-of-the-art visual explanation approaches: GradCAM \cite{selvaraju_grad-cam:_2017}, Gradient \cite{simonyan_deep_2014}, Integrated Gradient \cite{sundararajan_axiomatic_2017}, BBMP \cite{fong_interpretable_2017}, MGen \cite{dabkowski_real_2017} and SAGen \cite{Charachon2020CombiningSA}. Further details regarding these implementations are provided in \textbf{supp. section 5.2}.

\subsection{Assessing the quality of domain translation}\label{sec:metrics_DT}

Since there is no global consensus on how to measure the proximity between real and generated image distributions for domain translation, we revisit some methods and propose a novel evaluation. 
We propose to learn an embedding function independent from the visual explanation method that can separate in the latent space the distribution of real images predicted in class \textbf{0} from real images predicted in class \textbf{1}. This embedding is constructed by training a variational autoencoder (VAE) coupled with a multilayer perceptron that learns to classify if the mean encoded vector $\mu$ comes from an image in $\chi_0$ or $\chi_1$ \cite{Biffi2018LearningIA}. We use two different metrics to measure the distance between encoded distributions.
The first is inspired by the Fréchet Inception Distance \cite{Heusel2017GANsTB} where the Inception Network embedding (trained on natural images from ImageNet) is replaced by our VAE embedding in order to compute the Wasserstein-2 distance \cite{wasserstein1969markov}. It is denoted by $FD_{\mu}$. The second is based on kernel probability density estimation \cite{scott_skde} performed on a 2-dimensional Principal Component Analysis applied to the embedding VAE space. Real and generated densities are compared using Jenson-Shannon distance (JS) \cite{1207388}. 

% \end{document}

% \begin{document}

% \newcommand\f{{f_c}}
% \newcommand\thresh{\tau_{\f}}
% \newcommand\reals{\chi}
% \newcommand\E{{\chi_0}}
% \newcommand\F{{\chi_1}}
% \newcommand\cf{{c_{\f}}}
% \newcommand\gs{{g_{f_c}^{s\star}}}
% \newcommand\ga{{g_{f_c}^{a\star}}}
% \newcommand\gsopt{{g_{f_c}^{s}}}
% \newcommand\gaopt{{g_{f_c}^{a}}}
% \newcommand\Expl{{\cal E}}
% \newcommand\Fam{{{\cal F}_d}}

%----------------------------------------------------------------------------------------
%	Evaluation
%----------------------------------------------------------------------------------------

\section{Results}\label{sec:results}

\subsection{Evaluation of domain translation}\label{subsec:Domaintranslation}
\noindent\textbf{Classification Accuracy-} We measure the accuracy between the classifier's prediction on the original and the generated images to evaluate our method capacity to produce stable and adversarial images. Table \ref{tab:classif-performances} presents the classification accuracy of $\f$ predictions adversarial generated images against the original prediction $\cf(x)$. Results for stable generations are given in \textbf{suppl. sections 6.1}.
All the visual explanation methods trained with a classification target achieve to generate adversarial images classified in the opposite class. We observe that methods based on adversarial generation produce slightly better results (SAGen, CyCE and SyCE) compared to the perturbation mask approach (MGen) using Gaussian blur. For a similar architecture and training configuration, we also note that CyCE trained without classification target (CyCE w/o $L_{f_c}^{a, cy}$) i.e. a common CycleGAN, produces much poorer classification results.

\begin{table}[ht]
\caption{\textbf{Domain Translation results}. \textbf{(a)}: Accuracies computed between the original model's decision $c_{f_c}(x)$ and the decision on the adversarial image. \textbf{(b)} and \textbf{(c)}: Fréchet Distance ($FD_{\mu}$) and Jenson-Shannon distances ($JS$) on the different problems for respectively LeNet and ResNet50. A stands for Adversarial and \textit{St} for Stable generations.}
    \begin{subtable}[h]{\textwidth}
        \caption{Accuracy scores}
        \label{tab:classif-performances}
        \begin{adjustbox}{width=0.5\textwidth,center}
        \begin{small}
        \begin{sc}
        \begin{tabular}{lccc}
        \toprule
        Method &Digits& Pneumonia &Tumor Loc. \\
        \midrule
        Mgen&0.176&0.075&0.156 \\
        SAGen&0.032&0.103&0.243    \\
        CyCE w/o $L_{f_c}^{a, cy}$ &0.954&0.739&0.966\\
        CyCE &0.070&0.040&\textbf{0.090}\\
        SyCE&\textbf{0.015}&\textbf{0.028}&0.096   \\
        \bottomrule
        \end{tabular}
        \end{sc}
        \end{small}
        \end{adjustbox}
    \end{subtable}
    \begin{subtable}[h]{\textwidth}
    \begin{subtable}[h]{0.49\textwidth}
        \caption{$\chi_0 \longrightarrow \chi_1$}
        \label{tab:chi02chi1}
        \begin{adjustbox}{width=\textwidth,center}
        % \begin{center}
        \begin{small}
        \begin{sc}
        \begin{tabular}{lccccccc}
        \toprule
        Method && \multicolumn{2}{c}{Digits} & \multicolumn{2}{c}{Pneumonia} & \multicolumn{2}{c}{Tumor Loc.} \\
        && $FD_{\mu} \downarrow$& $JS \downarrow$&$FD_{\mu} \downarrow$& $JS \downarrow$&$FD_{\mu} \downarrow$& $JS \downarrow$ \\
        \midrule
        \midrule
        Mgen&A& 142.71& 0.99&58.76&0.87&65.29&0.43 \\
        \midrule
        \multirow{2}{*}{SAGen}&\textit{St}&\textit{(1.67)}& \textit{(0.60)}&\textit{(0.22)}& \textit{(0.09)}&\textit{(1.54)}& \textit{(0.14)}\\
        &A&55.93&  0.96 & 89.81&0.85&242.22&0.72\\
        \midrule
        CyCE&A&4.50&\textbf{0.55}&\textbf{1.92}&\textbf{0.31}&\textbf{2.66}&\textbf{0.29}\\
        \midrule
        % CycSAGen \\
        \multirow{2}{*}{SyCE}&\textit{St}&\textit{(0.30)}&\textit{(0.41)}&\textit{(0.04)}&\textit{(0.08)}&\textit{(0.13)}& \textit{(0.07)} \\
        &A&\textbf{2.84}& 0.58&2.12& 0.32&45.88&0.41\\
        % SAGen&5.59 \textit{(0.17)}&0.96 \textit{(0.6)}& 89.81 \textit{(0.22)}& 0.85 \textit{(0.09)}&\\
        % % \midrule
        % % CycSAGen \\
        % Ours&2.84 \textit{(0.03)}& 0.58 \textit{(0.41)}&2.12 \textit{(0.04)}&0.32 \textit{(0.08)}& \\
        \bottomrule
        \end{tabular}
        \end{sc}
        \end{small}
        \end{adjustbox}
    \end{subtable}
    \begin{subtable}[h]{0.49\textwidth}
        \caption{$\chi_1 \longrightarrow \chi_0$}
        \label{tab:chi12chi0}
        \begin{adjustbox}{width=\textwidth,center}
        % \begin{center}
        \begin{small}
        \begin{sc}
        \begin{tabular}{lccccccc}
        \toprule
        Method && \multicolumn{2}{c}{Digits} & \multicolumn{2}{c}{Pneumonia} & \multicolumn{2}{c}{Tumor Loc.} \\
        && $FD_{\mu} \downarrow$& $JS \downarrow$&$FD_{\mu} \downarrow$& $JS \downarrow$&$FD_{\mu}\downarrow$& $JS \downarrow$ \\
        \midrule
        \midrule
        Mgen& A&\textbf{3.10}& \textbf{0.60}&38.06&0.94&94.70&0.56\\
        \midrule
        \multirow{2}{*}{SAGen}&\textit{St}&\textit{(1.06)}& \textit{(0.40)}&\textit{(0.84)}& \textit{(0.33)}&\textit{(10.97)}&\textit{(0.10)}\\
        &A&12.04&  0.99 & 95.17&0.85&312.64&0.77\\
        \midrule
        CyCE&A&9.37& 0.75&\textbf{1.56}&\textbf{0.25}&\textbf{39.24}&\textbf{0.41}\\
        \midrule
        \multirow{2}{*}{SyCE}&\textit{St}&\textit{(0.30)}&\textit{(0.28)}&\textit{(0.36)}&\textit{(0.11)}&\textit{(0.05)}&\textit{(0.05)}\\
        &A&8.61& 0.71&9.71& 0.40&51.51&0.44\\
        \bottomrule
        \end{tabular}
        \end{sc}
        \end{small}
        \end{adjustbox}
    \end{subtable}
    \end{subtable}
\end{table}

\begin{figure}[h!]
\begin{subfigure}[h]{0.6\textwidth}
\begin{center}
\centerline{\includegraphics[width=\textwidth]{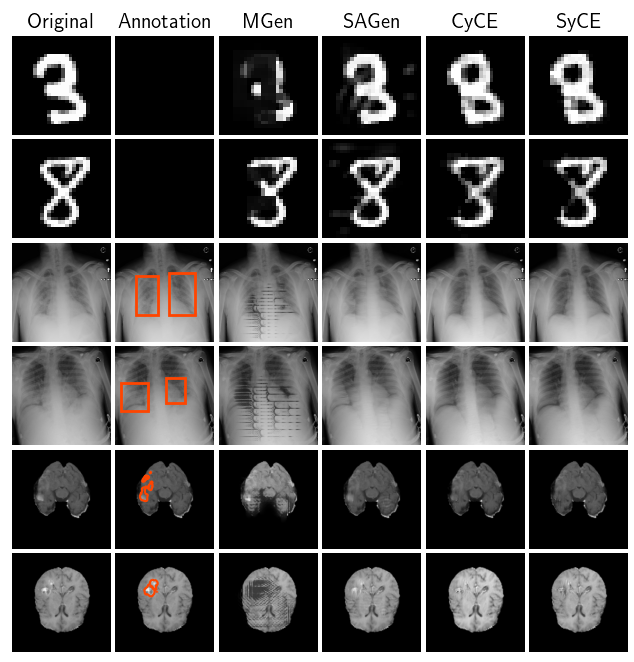}}
\caption{\textbf{Adversarial generations}}
% \caption{\textbf{Comparison of generated adversarial (or perturbed) images with other methods}. Given the original images and respective ground truth annotation (col. 1-2), we show the adversarial images generated by MGen, SAGen, CyCE and SyCE for the different tasks.}
\label{fig:dt_xr_br}
\end{center}
\end{subfigure}
\begin{subfigure}[h]{0.38\textwidth}
    \begin{subfigure}[h]{\textwidth}
	\begin{center}
		\centerline{\includegraphics[width=\textwidth]{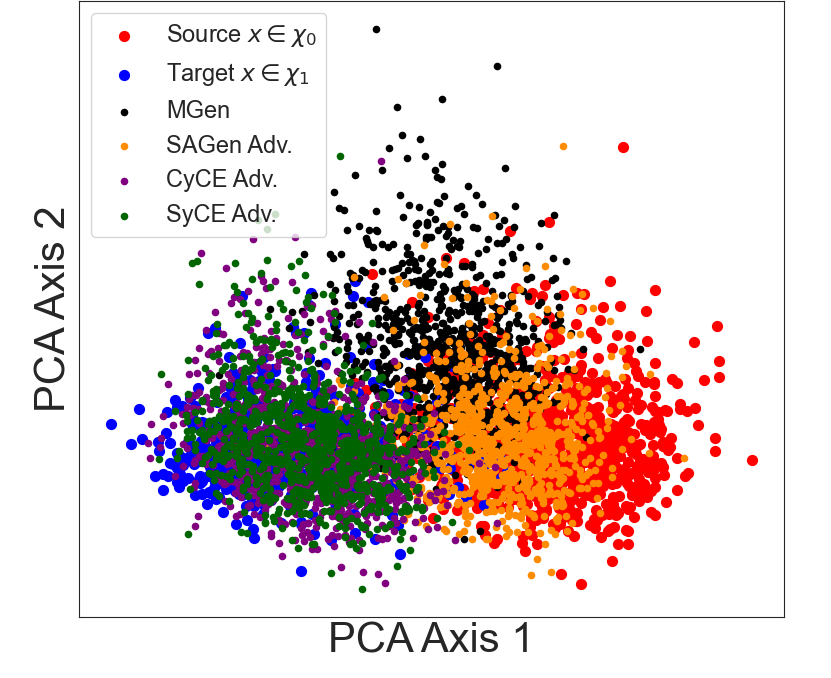}}
		\caption{$\chi_0 \longrightarrow \chi_1$}
		\label{fig:pca_xrays_01_adv}
	\end{center}
    \end{subfigure}
    \begin{subfigure}[h]{\textwidth}
    	\begin{center}
    		\centerline{\includegraphics[width=\textwidth]{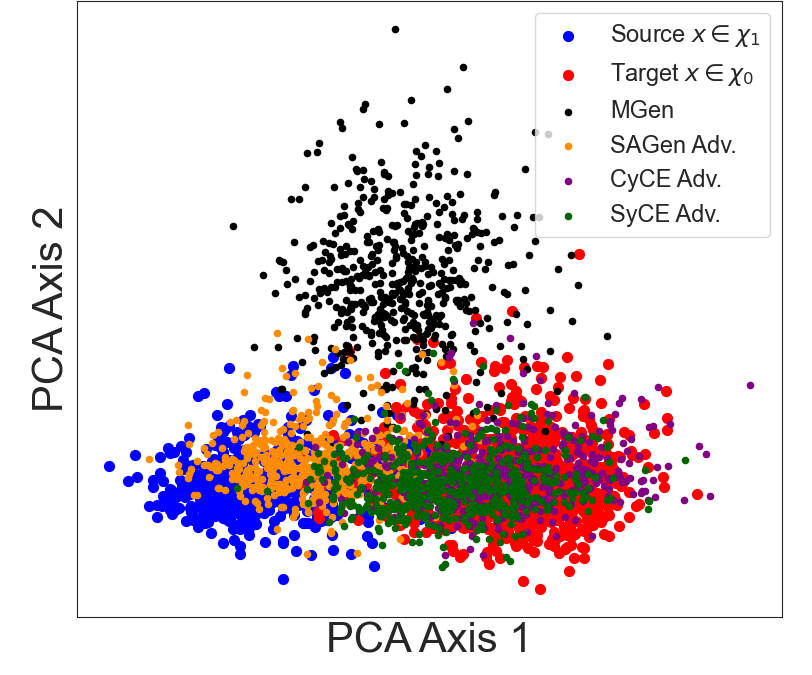}}
    		\caption{$\chi_1 \longrightarrow \chi_0$}
    		\label{fig:pca_xrays_10_adv}
    	\end{center}
    \end{subfigure}
    \end{subfigure}
	\caption{\textbf{Qualitative Results}. \textbf{(a)}: Comparison of generated adversarial (or perturbed) images with other methods. \textbf{(b)} and \textbf{(c)}: First 2 axes of the PCA applied on the embedded vector $\mu$ of the VAE for all images (real and generated) of the test set for \textbf{Pneumonia detection}. \textbf{(b)}: Source (original) domain $\chi_0$ and Target for adversarial generation: $\chi_1$. \textbf{(c)}: Source $\chi_1$ and Target $\chi_0$.}
\end{figure}

\noindent\textbf{Domain Translation Quality-} We recall the "embedding" metrics introduced in \ref{sec:metrics_DT}: Fréchet Distance on the mean encoded vector ($FD_{\mu}$) and the Jenson-Shannon distance ($JS$) on the estimated 2-dimensional distribution.\\
Figure \ref{fig:dt_xr_br} displays some examples of adversarial generated images for different explanation approaches using a generation process (only outputs of generations transforming  pathological to healthy for medical imaging tasks). We point out several observations: 
(i) Except for digits in direction "8" to "3", MGen generates adversaries that humans perceive as synthetic. 
(ii) SAGen produces adversarial images that are actually very similar to the original images, there is no domain transfer for human eyes. 
(iii) Our approaches (CyCE and SyCE) generate adversarial images which are perceived as real images of the opposite domain e.g. in MR, bright  focal regions (tumors) are replaced by darker regions in the adversaries (healthy tissue). 
These visual findings are supported by the results shown in tables \ref{tab:chi12chi0} and \ref{tab:chi02chi1}, as well as the PCA representations in Figures \ref{fig:pca_xrays_01_adv} and \ref{fig:pca_xrays_10_adv}. 
SyCE and CyCE significantly outperform both MGen and SAGen in producing adversaries closer to the opposite image distribution. 
% (except again for MGen on Digits when translating from $\chi_1$ to $\chi_0$).
SAGen sometimes has the lowest performance despite a higher perception of realism than MGen.
We observe in Figures \ref{fig:pca_xrays_01_adv} and \ref{fig:pca_xrays_10_adv} (for pneumonia) that SAGen generates images (orange points) that remain very close to the original distribution 
% (red in Fig. \ref{fig:pca_xrays_v3_01} and blue in Fig. \ref{fig:pca_xrays_v3_10})
while those produced by MGen (black) are often far from both real distributions. % $\chi_0$ and $\chi_1$. 
% (but could be closer to the opposite distribution). 
CyCE performs best in most cases in both directions (although SyCE remains competitive). Compared to the cyclic constraint in CyCE, the symmetry (in SyCE) is more restrictive and better enforces the generated adversarial images to be close to the original image. Finally, we see in tables \ref{tab:chi12chi0} and \ref{tab:chi02chi1} that stable images (St) produced by our method (SyCE) are often slightly closer to the real distribution of the original image compared to the SAGen stable generation. See \textbf{suppl. sections 6.1} and \textbf{7} for additional figures and results.

\noindent\textbf{Limitations- }CyCE and SyCE have good performances in capturing the different types and locations of impactful regions for the classifier when acting on pathological images. Yet, they fail to capture pathology variability when mapping healthy to pathological images. 
Healthy images are often perturbed in similar locations. This is a known drawback of symmetric and cyclic constraints forcing a one-to-one map between domains of different complexity \cite{bashkirova2019adversarial}. \\

\begin{figure}[h!]
\begin{center}
\centerline{\includegraphics[width=0.75\textwidth]{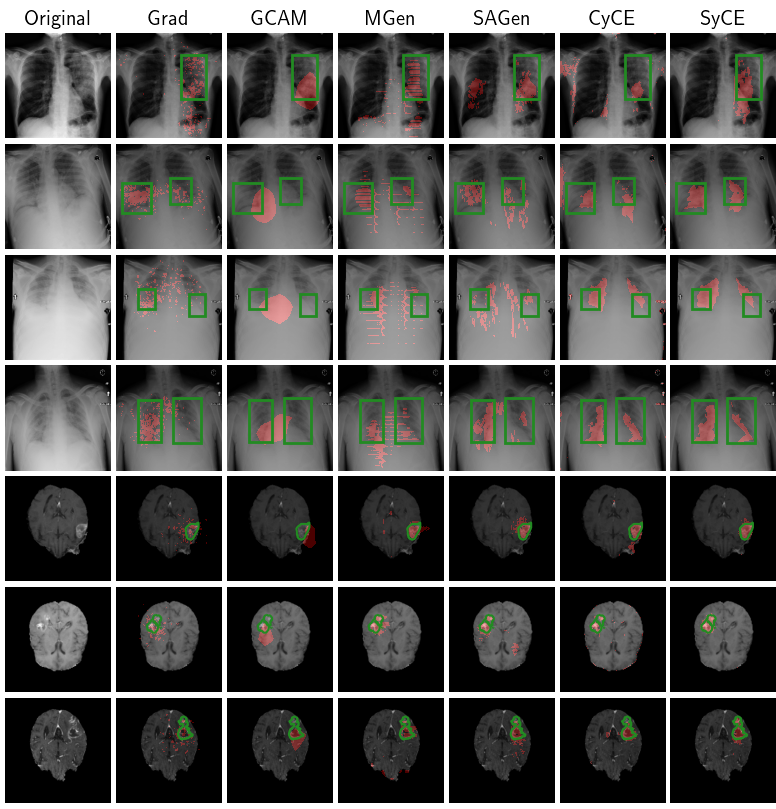}}
% \caption{\textbf{Comparison with features attribution techniques and against ground truth annotations}. Top lines: Pneumonia detection on X-Rays. Bottom lines: Brain MRI tumor localization.}
\caption{\textbf{Comparison with features attribution techniques and against ground truth annotations}. Top lines: Pneumonia detection on X-rays. All explanation maps are thresholded at the 95th percentile and binarized. Bottom lines: Brain Tumor localization (Threshold at the 98th perc.).}
\label{fig:heatmaps_xr_br_paper}
\end{center}
\end{figure}

\subsection{Relevant regions found by the visual explanation}\label{subsec:FA}
\noindent\textbf{Localization Metrics- } A common method to evaluate visual explanation techniques is to correlate the feature attributions they produce with human annotations. For a competitive classifier, we expect the highlighted supporting regions of the input image to match human annotations.

\begin{table}[ht]
\caption{\textbf{Localization results}. Results for ResNet50 and DenseNet121  on Pneumonia detection and Brain tumor localization problems.}
% \caption{\textbf{Localization results}. Results are given on Pneumonia detection and Brain tumor localization problems for visual explanations of respectively ResNet50 and DenseNet121 classifiers.}
\begin{subtable}[h]{0.48\textwidth}
\caption{ResNet50}
\label{tab:loc_results_resnet50}
\begin{adjustbox}{width=\textwidth,center}
% \begin{center}
\begin{small}
\begin{sc}
\begin{tabular}{lcccccccc}
\toprule
Method & \multicolumn{4}{c}{Pneumonia} & &\multicolumn{3}{c}{Tumor Loc.} \\
& $IoU_{90}$& $IoU_{95}$ & $IoU_{98}$ & NCC && $IoU_{98}$ & $IoU_{99}$ & NCC \\
\midrule
Gradient&0.187&0.152&0.097&0.312&&0.154&0.131&0.330\\
IG& 0.170&0.136&0.086&0.254&&0.238&0.196&0.444\\
GCAM& 0.195&0.138&0.070&0.325&&0.173&0.115&0.389\\
BBMP&0.204&0.154&0.087& 0.348&&0.290&0.263&0.409\\
Mgen&0.208&0.169&0.103& 0.340&&0.319&0.274&0.448\\
SAGen & 0.232&0.173&0.097&0.325&&0.330&0.284&0.515 \\
CyCE&0.221&0.191&0.116&0.337&&0.322&0.270&0.516 \\
% &CycSAGen & 0.291&0.238&0.146&0.461&&0.240&0.271&0.257&0.424\\
SyCE w/o St.&0.292&0.236&0.142&0.494&&0.406&0.345&0.609\\
SyCE & \textbf{0.299}&\textbf{0.244}&\textbf{0.151}&  \textbf{0.506}&&\textbf{0.411}&\textbf{0.348}&\textbf{0.615}\\
\bottomrule
\end{tabular}
\end{sc}
\end{small}
% \end{center}
\end{adjustbox}
\vskip -0.25cm
\end{subtable}
\begin{subtable}[h]{0.48\textwidth}
\caption{DenseNet121}
\label{tab:loc_results_densenet121}
\begin{adjustbox}{width=\textwidth,center}
% \begin{center}
\begin{small}
\begin{sc}
\begin{tabular}{lcccccccc}
\toprule
Method & \multicolumn{4}{c}{Pneumonia} & &\multicolumn{3}{c}{Tumor Loc.} \\
& $IoU_{90}$& $IoU_{95}$ & $IoU_{98}$ & NCC && $IoU_{98}$ & $IoU_{99}$ & NCC \\
\midrule
Gradient&0.159&0.127&0.081&0.267&&0.128&0.099&0.261\\
IG&0.123&0.095&0.074&0.181&&0.206&0.168&0.397\\
GCAM&0.223&0.174&0.085&0.344&&0.220&0.111&0.342\\
Mgen&0.264&0.202&0.105&0.338&&0.333&0.284&0.519\\
SAGen&0.255&0.191&0.107&0.337&&0.264&0.222& 0.440  \\
CyCE &0.251&0.205&0.111&0.393&&0.264&0.236&0.424\\
SyCE w/o St.&0.271&0.221&0.130&0.428&&0.350&0.310&0.558\\
SyCE &\textbf{0.284}&\textbf{0.235}&\textbf{0.144}&\textbf{0.460}&&\textbf{0.372}&\textbf{0.329}&\textbf{ 0.582}  \\
\bottomrule
\end{tabular}
\end{sc}
\end{small}
% \end{center}
\end{adjustbox}
\vskip -0.2cm
\end{subtable}
\end{table}

% We consider two metrics to evaluate the localization performance of visual explanation techniques: 
To evaluate this localization performance, we consider:
(i) the intersection over union ($IoU$) computed between binary ground truth annotation and a thresholded binary explanation mask.
The choice of the thresholds depends on the representative size of the annotations on the training set. For instance, if human annotations occupy in average 10\% of the image, the threshold is set at the 90th percentile of the explanation map ($IOU_{90}$) in tables \ref{tab:loc_results_resnet50} and  \ref{tab:loc_results_densenet121}). (ii) The normalized cross-correlation ($NCC$) directly computed between the binary ground truth mask and the raw visual explanation as it is not sensitive to the intensity (see \textbf{suppl. section 6.2} for further details).

\noindent\textbf{Visual and Quantitative Evaluations- }Tables \ref{tab:loc_results_resnet50} and \ref{tab:loc_results_densenet121} show the results for these two metrics comparing our work against state-of-the-art approaches. Two trained classifier, for both Pneumonia detection and Brain tumor localization are evaluated (ResNet-50 and DenseNet-121).
Our method (SyCE) outperforms all others for the two problems and for the two classifiers. Some binary visual explanations are shown in Figure \ref{fig:heatmaps_xr_br_paper}. SyCE produces explanation maps more attached to the image structures while pointing out different supporting regions that are human-understandable. Moreover, the contribution of the stable image is less important than what is described in \cite{Charachon2020CombiningSA}, even if it slightly improves the localization (see tables \ref{tab:loc_results_resnet50} and \ref{tab:loc_results_densenet121}) \textsc{SyCE w/o St.} vs \textsc{SyCE}). This is due to a generation process which is not penalized with $L_p$ norms as in \cite{Charachon2020CombiningSA,Elliott2019AdversarialPO}.
Finally, although CyCE is the best performer for domain translation (see section \ref{subsec:Domaintranslation}) and is competitive with other works of the literature, it obtains poorer localization results than SyCE. Figure \ref{fig:heatmaps_xr_br_paper} shows that CyCE produces more irrelevant attributions, underlining the benefit of symmetrical constraints in SyCE.
Compared to CyCE, SyCE perturbs less regions. Perturbed regions by SyCE are more discriminating for the classifier.
% Generators in SyCE perturb less regions, but regions that are more discriminating for the classifier compared to CyCE.
% Generators in SyCE perturb less the most discriminating regions than CyCE.
Additional figures and results are provided in \textbf{suppl. sections 6.2} and \textbf{7}.

% \end{document}

% \begin{document}

%----------------------------------------------------------------------------------------
%	Conclusion
%----------------------------------------------------------------------------------------
\section{Conclusion}\label{sec:conclusion}

We have introduced a general formulation to design a visual explanation of classifier decisions. Our method captures important patterns in the input image on which the classifier relies to make its decision. We propose one implementation that strictly follows the general formulation as well as a weaker and relaxed version. Both leverage generative image-to-image translation frameworks, cycle consistency or even symmetry constraints. Compared to previous works, we show on different datasets that our symmetrical method better localizes discriminative regions for the classifier that are interpretable by humans, in particular for clinicians in the medical domain. Through a revisited use of variational autoencoders, we have successfully validated that our techniques are able to either stabilize or transpose an image to respectively its original or counterfactual image distribution, while ensuring the proximity to the input image.

% \end{document}

\bibliographystyle{splncs04}
\bibliography{bibliography}

\begin{thebibliography}{10}
\providecommand{\url}[1]{\texttt{#1}}
\providecommand{\urlprefix}{URL }
\providecommand{\doi}[1]{https://doi.org/#1}

\bibitem{adebayo_sanity_2018}
Adebayo, J., Gilmer, J., Muelly, M., Goodfellow, I.J., Hardt, M., Kim, B.:
  Sanity checks for saliency maps. In: NeurIPS (2018)

\bibitem{bashkirova2019adversarial}
Bashkirova, D., Usman, B., Saenko, K.: Adversarial self-defense for
  cycle-consistent gans. In: NeurIPS (2019)

\bibitem{Bass2020ICAMIC}
Bass, C., da~Silva, M., Sudre, C., Tudosiu, P.D., Smith, S., Robinson, E.:
  Icam: Interpretable classification via disentangled representations and
  feature attribution mapping. In: NeurIPS (2020)

\bibitem{Baumgartner2018VisualFA}
Baumgartner, C.F., Koch, L., Tezcan, K.C., Ang, J.X., Konukoglu, E.: Visual
  feature attribution using wasserstein gans. In: CVPR (2018)

\bibitem{mrnet}
Bien, N., Rajpurkar, P., Ball, R., Irvin, J., Park, A., Jones, E., Bereket, M.,
  Patel, B., Yeom, K., Shpanskaya, K., Halabi, S., Zucker, E., Fanton, G.,
  Amanatullah, D., Beaulieu, C., Riley, G., Stewart, R., Blankenberg, F.,
  Larson, D., Lungren, M.: Deep-learning-assisted diagnosis for knee magnetic
  resonance imaging: Development and retrospective validation of mrnet. PLOS
  Medicine  \textbf{15} (2018)

\bibitem{Biffi2018LearningIA}
Biffi, C., Oktay, O., Tarroni, G., Bai, W., Marvao, A.S.M.D., Doumou, G.,
  Rajchl, M., Bedair, R., Prasad, S.K., Cook, S., O’Regan, D., Rueckert, D.:
  Learning interpretable anatomical features through deep generative models:
  Application to cardiac remodeling. In: MICCAI (2018)

\bibitem{chang_explaining_2019}
Chang, C.H., Creager, E., Goldenberg, A., Duvenaud, D.K.: Explaining image
  classifiers by counterfactual generation. In: ICLR (2019)

\bibitem{Charachon2020CombiningSA}
Charachon, M., Hudelot, C., Courn{\`e}de, P.H., Ruppli, C., Ardon, R.:
  Combining similarity and adversarial learning to generate visual explanation:
  Application to medical image classification. In: ICPR (2020)

\bibitem{dabkowski_real_2017}
Dabkowski, P., Gal, Y.: Real time image saliency for black box classifiers. In:
  NIPS (2017)

\bibitem{imagenet_cvpr09}
Deng, J., Dong, W., Socher, R., Li, L.J., Li, K., Fei-Fei, L.: {ImageNet: A
  Large-Scale Hierarchical Image Database}. In: CVPR (2009)

\bibitem{dhurandhar_explanations_2018}
Dhurandhar, A., Chen, P.Y., Luss, R., Tu, C.C., Ting, P.S., Shanmugam, K., Das,
  P.: Explanations based on the missing: Towards contrastive explanations with
  pertinent negatives. In: NeurIPS (2018)

\bibitem{Elliott2019AdversarialPO}
Elliott, A., Law, S., Russell, C.: Adversarial perturbations on the perceptual
  ball. ArXiv  \textbf{abs/1912.09405} (2019)

\bibitem{1207388}
{Endres}, D.M., {Schindelin}, J.E.: A new metric for probability distributions.
  IEEE Transactions on Information Theory  \textbf{49}(7) (2003)

\bibitem{skincancerEsteva2017}
Esteva, A., Kuprel, B., Novoa, R., Ko, J., Swetter, S., Blau, H., Thrun, S.:
  Dermatologist-level classification of skin cancer with deep neural networks.
  In: Nature. vol.~542 (2017)

\bibitem{fong_interpretable_2017}
{Fong}, R.C., {Vedaldi}, A.: Interpretable explanations of black boxes by
  meaningful perturbation. In: ICCV (2017)

\bibitem{Fong2019UnderstandingDN}
Fong, R., Patrick, M., Vedaldi, A.: Understanding deep networks via extremal
  perturbations and smooth masks. In: ICCV (2019)

\bibitem{NIPS2014_5423}
Goodfellow, I., Pouget-Abadie, J., Mirza, M., Xu, B., Warde-Farley, D., Ozair,
  S., Courville, A., Bengio, Y.: Generative adversarial nets. In: NIPS (2014)

\bibitem{Goyal2019CounterfactualVE}
Goyal, Y., Wu, Z., Ernst, J., Batra, D., Parikh, D., Lee, S.: Counterfactual
  visual explanations. In: ICML (2019)

\bibitem{Gulrajani2017ImprovedTO}
Gulrajani, I., Ahmed, F., Arjovsky, M., Dumoulin, V., Courville, A.C.: Improved
  training of wasserstein gans. In: NIPS (2017)

\bibitem{DBLP:journals/corr/HeZR016}
He, K., Zhang, X., Ren, S., Sun, J.: Identity mappings in deep residual
  networks. In: ECCV (2016)

\bibitem{Heusel2017GANsTB}
Heusel, M., Ramsauer, H., Unterthiner, T., Nessler, B., Hochreiter, S.: Gans
  trained by a two time-scale update rule converge to a local nash equilibrium.
  In: NIPS (2017)

\bibitem{hsieh2020evaluations}
Hsieh, C.Y., Yeh, C.K., Liu, X., Ravikumar, P., Kim, S., Kumar, S., Hsieh,
  C.J.: Evaluations and methods for explanation through robustness analysis.
  ArXiv  \textbf{abs/2006.00442} (2020)

\bibitem{Huang2017DenselyCC}
Huang, G., Liu, Z., Weinberger, K.Q.: Densely connected convolutional networks.
  In: CVPR (2017)

\bibitem{pix2pix2017}
Isola, P., Zhu, J.Y., Zhou, T., Efros, A.A.: Image-to-image translation with
  conditional adversarial networks. In: CVPR (2017)

\bibitem{DBLP:journals/corr/KingmaB14}
Kingma, D.P., Ba, J.: Adam: {A} method for stochastic optimization. In: ICLR
  (2015)

\bibitem{lenet}
{Lecun}, Y., {Bottou}, L., {Bengio}, Y., {Haffner}, P.: Gradient-based learning
  applied to document recognition. Proceedings of the IEEE  \textbf{86}(11)
  (1998)

\bibitem{lecun-mnisthandwrittendigit-2010}
LeCun, Y., Cortes, C.: {MNIST} handwritten digit database  (2010),
  \url{http://yann.lecun.com/exdb/mnist/}

\bibitem{Lenis2020DomainAM}
Lenis, D., Major, D., Wimmer, M., Berg, A., Sluiter, G., B{\"u}hler, K.: Domain
  aware medical image classifier interpretation by counterfactual impact
  analysis. In: MICCAI (2020)

\bibitem{Liu2017UnsupervisedIT}
Liu, M.Y., Breuel, T., Kautz, J.: Unsupervised image-to-image translation
  networks. In: NIPS (2017)

\bibitem{major_interpreting_2020}
Major, D., Lenis, D., Wimmer, M., Sluiter, G., Berg, A., B{\"u}hler, K.:
  Interpreting medical image classifiers by optimization based counterfactual
  impact analysis. In: ISBI (2020)

\bibitem{Mirza2014ConditionalGA}
Mirza, M., Osindero, S.: Conditional generative adversarial nets. ArXiv
  \textbf{abs/1411.1784} (2014)

\bibitem{Narayanaswamy2020ScientificDB}
Narayanaswamy, A., Venugopalan, S., Webster, D.R., Peng, L., Corrado, G.,
  Ruamviboonsuk, P., Bavishi, P., Brenner, M., Nelson, P., Varadarajan, A.V.:
  Scientific discovery by generating counterfactuals using image translation.
  In: MICCAI (2020)

\bibitem{DBLP:journals/corr/RonnebergerFB15}
Ronneberger, O., Fischer, P., Brox, T.: U-net: Convolutional networks for
  biomedical image segmentation. In: MICCAI (2015)

\bibitem{scott_skde}
{Scott}, D.W.: Multivariate density estimation: Theory, practice, and
  visualization. John Wiley \& Sons  (1992)

\bibitem{seah_chest_2019}
Seah, J.C.Y., Tang, J.S.N., Kitchen, A., Gaillard, F., Dixon, A.F.: Chest
  {Radiographs} in {Congestive} {Heart} {Failure}: {Visualizing} {Neural}
  {Network} {Learning}. In: Radiology. vol.~290

\bibitem{selvaraju_grad-cam:_2017}
Selvaraju, R.R., Cogswell, M., Das, A., Vedantam, R., Parikh, D., Batra, D.:
  Grad-{CAM}: {Visual} {Explanations} from {Deep} {Networks} via
  {Gradient}-{Based} {Localization}. In: {ICCV} (2017)

\bibitem{shen_one--one_2020}
Shen, Z., Chen, Y., Huang, T.S., Zhou, S., Georgescu, B., Liu, X.: One-to-one
  {Mapping} for {Unpaired} {Image}-to-image {Translation}. In: WACV (2020)

\bibitem{simonyan_deep_2014}
Simonyan, K., Vedaldi, A., Zisserman, A.: Deep {Inside} {Convolutional}
  {Networks}: {Visualising} {Image} {Classification} {Models} and {Saliency}
  {Maps}. In: ICLR (2014)

\bibitem{Simpson2019ALA}
Simpson, A., Antonelli, M., Bakas, S., Bilello, M., Farahani, K., Ginneken, B.,
  Kopp-Schneider, A., Landman, B., Litjens, G., Menze, B., Ronneberger, O.,
  Summers, R., Bilic, P., Christ, P., Do, R., Gollub, M., Golia-Pernicka, J.,
  Heckers, S., Jarnagin, W., McHugo, M., Napel, S., Vorontsov, E., Maier-Hein,
  L., Cardoso, M.J.: A large annotated medical image dataset for the
  development and evaluation of segmentation algorithms. ArXiv
  \textbf{abs/1902.09063} (2019)

\bibitem{Singla2020ExplanationBP}
Singla, S., Pollack, B., Chen, J., Batmanghelich, K.: Explanation by
  progressive exaggeration. In: ICLR (2020)

\bibitem{smilkov_smoothgrad:_2017}
Smilkov, D., Thorat, N., Kim, B., Vi{\'e}gas, F.B., Wattenberg, M.: Smoothgrad:
  removing noise by adding noise. ArXiv  \textbf{abs/1706.03825} (2017)

\bibitem{springenberg_striving_2014}
Springenberg, J.T., Dosovitskiy, A., Brox, T., Riedmiller, M.A.: Striving for
  simplicity: The all convolutional net. In: ICLR. vol. abs/1412.6806 (2015)

\bibitem{sundararajan_axiomatic_2017}
Sundararajan, M., Taly, A., Yan, Q.: Axiomatic attribution for deep networks.
  In: ICML (2017)

\bibitem{Wang2020SCOUTSD}
Wang, P., Vasconcelos, N.: Scout: Self-aware discriminant counterfactual
  explanations. In: CVPR (2020)

\bibitem{DBLP:journals/corr/WangPLLBS17}
Wang, X., Peng, Y., Lu, L., Lu, Z., Bagheri, M., Summers, R.M.: Chestx-ray8:
  Hospital-scale chest x-ray database and benchmarks on weakly-supervised
  classification and localization of common thorax diseases. In: CVPR (2017)

\bibitem{wasserstein1969markov}
Wasserstein, L.N.: Markov processes over denumerable products of spaces
  describing large systems of automata. Problems of Information Transmission
  \textbf{5}(3) (1969)

\bibitem{Wolleb2020DeScarGANDA}
Wolleb, J., Sandk{\"u}hler, R., Cattin, P.: Descargan: Disease-specific anomaly
  detection with weak supervision. In: MICCAI (2020)

\bibitem{woods_adversarial_2019}
Woods, W., Chen, J., Teuscher, C.: Adversarial explanations for understanding
  image classification decisions and improved neural network robustness. In:
  Nature Machine Intelligence. vol.~1 (2019)

\bibitem{Zhou2016LearningDF}
Zhou, B., Khosla, A., Lapedriza, {\`A}., Oliva, A., Torralba, A.: Learning deep
  features for discriminative localization. In: CVPR (2016)

\bibitem{CycleGAN2017}
Zhu, J.Y., Park, T., Isola, P., Efros, A.A.: Unpaired image-to-image
  translation using cycle-consistent adversarial networks. In: ICCV (2017)

\end{thebibliography}

\appendix
% This is samplepaper.tex, a sample chapter demonstrating the
% LLNCS macro package for Springer Computer Science proceedings;
% Version 2.20 of 2017/10/04
%
% \documentclass[runningheads]{llncs}
% %
% \usepackage{graphicx}
% \graphicspath{{images/}}
% \usepackage{subfiles}
% \usepackage{multirow}
% \usepackage{xspace}
% \usepackage{amsmath}
% \usepackage{amssymb}
% \usepackage{adjustbox}
% \usepackage{caption}
% \usepackage{subcaption}
% \usepackage{booktabs}
% \usepackage{bbding}
% \usepackage{color}
% \usepackage{hyperref}
% \begin{document}

% Our code is available on github at
% \url{https://github.com/martinCharachon/cGANsExplainer}
% \url{https://anonymous.4open.science/r/26e23a4f-14a0-43f6-9287-979847e2851a/}
\title{Supplementary Materials: \\ Leveraging Conditional Generative Models \\ in a General Explanation Framework of Classifier Decisions}
\titlerunning{Supplementary Materials}

% \author{Martin Charachon\inst{1, 2} \and
% Paul-Henry Cournède\inst{2} \and
% Céline Hudelot\inst{2} \and 
% Roberto Ardon\inst{1}}
% \authorrunning{M. Charachon et al.}
% \institute{Incepto Medical, France \and
% MICS, Université Paris-Saclay, CentraleSupélec, France}

\maketitle   
% \documentclass[../main.tex]{subfiles}
% \graphicspath{{\subfix{images/}}}

% \begin{document}

% %----------------------------------------------------------------------------------------
% %	Reminder
% %----------------------------------------------------------------------------------------

% \newcommand\f{{f_c}}
% \newcommand\thresh{\tau_{\f}}
% \newcommand\reals{\chi}
% \newcommand\E{{\chi_0}}
% \newcommand\F{{\chi_1}}
% \newcommand\cf{{c_{\f}}}
% \newcommand\gs{{g_{f_c}^{s\star}}}
% \newcommand\ga{{g_{f_c}^{a\star}}}
% \newcommand\gsopt{{g_{f_c}^{s}}}
% \newcommand\gaopt{{g_{f_c}^{a}}}
% \newcommand\Expl{{\cal E}}
% \newcommand\gz{{g_0^{\star}}}
% \newcommand\gu{{g_1^{\star}}}

\section*{\textcolor{red}{Reminder: General Formulation}}
\textbf{Visual explanation definition- }
\begin{equation}\label{eq:ExplanationDef_supp}
	\Expl(x) = |\gs(x) - \ga(x)|
\end{equation}

\textbf{General optimization problem- }
\begin{equation}\label{eq:paperformulation_supp}	
	(\gs, \ga) = \hspace{-1.4cm}
	\underset{
			\begin{array}{c}
				\scriptstyle \gsopt, \gaopt \\
				\scriptstyle \\
			    \underbrace{
				\scriptstyle \text{s.t. } 
				\left\{		
				\hspace{-1mm}
				\begin{array}{c}
					\scriptstyle \gsopt(\chi_0) \subset \chi_0, \gsopt(\chi_1) \subset \chi_1 \\
					\scriptstyle \gaopt(\chi_0) \subset \chi_1, \gaopt(\chi_1) \subset \chi_0
				\end{array}
				\hspace{-1mm}
				\right\} }_{\scriptstyle \text{\color{blue} \textbf{Consistency with reality}}}
			\end{array} }
		{\mathrm{argmin}} \hspace{-1.3cm}	
		\Big(
		\overbrace{
		\small
		\underbrace{d_g(\gsopt, \gaopt)}_{
		\text{\color{blue} \textbf{Regularity} }} +
		\mathbb{E}_x   \left (
		\hspace{-1mm} + \hspace{-1mm} 
		\begin{array}{c}
			d_s(x, \gsopt(x)) \\ 
			d_a(x, \gaopt(x)) 
		\end{array}
		\hspace{-1mm}
		\right )
		}^{\text{\color{blue} \textbf{Relevance}}} 	
		\Big)
\end{equation}

\section*{\textcolor{red}{Reminder: Proposed embodiments}}
\subsection*{SyCE: Symmetrically Conditioned Explanation}
\textbf{Generators definition- }Introducing auxialiary generators $g_0$ and $g_1$ (see the main paper), we define $\ga$ and $\gs$ as follow:
\begin{equation}\label{eq:sym_adversary_supp}
	\ga(x) = \left \{
	\begin{array}{l}
		\gz(x) \in \chi_1 \text{ if } x\in\chi_0 \\
	    \gu(x) \in \chi_0 \text{ if } x\in\chi_1 \\
	\end{array}\hspace{-0.1cm} 
	\right.
\end{equation}

\begin{equation}\label{eq:sym_similar_supp}
	\gs(x) = \left \{
	\begin{array}{l}
		\gz^2(x) \in \chi_0 \text{ if } x\in\chi_0\\
		\gu^2(x) \in \chi_1 \text{ if } x\in\chi_1\\
	\end{array}\hspace{-0.1cm} 
	\right.
\end{equation}

\noindent\textbf{Visual explanation- }
\begin{equation}\label{eq:ExplanationDefg0g1SyCE_supp}
	\Expl(x) = 
	\left \{\begin{array}{cc}
		|\gz^2(x) - \gz(x)|\, \text{ if }\,  x \in \chi_0 \\
		|\gu^2(x) - \gu(x)|\, \text{ if }\,  x \in \chi_1
	\end{array}
	\right \}
\end{equation}

\noindent\textbf{Optimization problem- }Equation \eqref{eq:SyCE_supp} is the proposed approximated optimization problem for SyCE. 
\begin{equation}\label{eq:SyCE_supp}
(\gz, \gu ) = \hspace{-0.9cm}
	\underset{
		\begin{array}{c}
			\scriptstyle g_0, g_1 \\
			\scriptstyle \text{s.t.} \\
			\scriptstyle g_0(\chi_0) \subset \chi_1, g_1(\chi_1) \subset \chi_0\\
			\scriptstyle g_0^2(\chi_0) \subset \chi_0, g_1^2(\chi_1) \subset \chi_1
		\end{array}
	}{\mathrm{argmin}} \hspace{-0.6cm}
	\left[  
	\begin{array}{c}
		\mathbb{E}_{x \in \chi_0} \left (||x - g_0^2(x)|| \right )  \\  + \\
		\mathbb{E}_{x \in \chi_1} \left (||x - g_1^2(x)||\right )
	\end{array}
	 \right]
\end{equation}

\subsection*{CyCE: Cyclic Conditioned Explanation}
\textbf{Generators definition- }As presented in the main paper, CyCE is a more relaxed approximation of problem \eqref{eq:paperformulation_supp}, where auxiliary generators $g_0$ and $g_1$ are introduced such that:
\begin{equation}\label{eq:sym_adversary_supp}
	\ga(x) = \left \{
	\begin{array}{l}
		\gz(x) \in \chi_1 \text{ if } x\in\chi_0 \\
	    \gu(x) \in \chi_0 \text{ if } x\in\chi_1 \\
	\end{array}\hspace{-0.1cm} 
	\right.
\end{equation}
And
\begin{equation}\label{eq:sym_similar_supp}
	\gs(x) = x, \, \, \, \forall x\in \chi
\end{equation}
\noindent\textbf{Visual explanation- }
\begin{equation}\label{eq:ExplanationDefg0g1CyCE_supp}
	\Expl(x) = 
	\left \{\begin{array}{cc}
		|x - \gz(x)|\, \text{ if }\,  x \in \chi_0 \\
		|x - \gu(x)|\, \text{ if }\,  x \in \chi_1
	\end{array}
	\right \}
\end{equation}

\noindent\textbf{Optimization problem- }Leveraging cycle consistency contraint, the approximated problem reads

\begin{equation}\label{eq:CyCE_supp}
	(\gz, \gu ) = \hspace{-0.9cm}
	\underset{
		\begin{array}{c}
			\scriptstyle g_0, g_1\\
			\scriptstyle \text{s.t.} \\
			\scriptstyle g_0(\chi_0) \subset \chi_1, g_1(\chi_1) \subset \chi_0
		\end{array}
	}{\mathrm{argmin}} \hspace{-0.6cm}
	\left[  
	\small
	\begin{array}{c}
		\mathbb{E}_{x \in \chi_0} \left (||x -  g_1(g_0(x))|| \right )  \\  + \\
		\mathbb{E}_{x \in \chi_1} \left (||x -  g_0(g_1(x))||\right )
	\end{array}
	 \right]
\end{equation}

% \end{document}
% \documentclass[../main.tex]{subfiles}
% \graphicspath{{\subfix{images/}}}

% \begin{document}

% \newcommand\f{{f_c}}
% \newcommand\thresh{\tau_{\f}}
% \newcommand\reals{\chi}
% \newcommand\E{{\chi_0}}
% \newcommand\F{{\chi_1}}
% \newcommand\cf{{c_{\f}}}
% \newcommand\gs{{g_{f_c}^{s\star}}}
% \newcommand\ga{{g_{f_c}^{a\star}}}
% \newcommand\gsopt{{g_{f_c}^{s}}}
% \newcommand\gaopt{{g_{f_c}^{a}}}
% \newcommand\Expl{{\cal E}}
% \newcommand\gz{{g_0^{\star}}}
% \newcommand\gu{{g_1^{\star}}}
%----------------------------------------------------------------------------------------
%	Other possible embodiments
%----------------------------------------------------------------------------------------
\section{Other possible embodiments of problem \eqref{eq:paperformulation_supp}}\label{sec:otherembodiment}
\subsection{SSyE: Strict Symmetrical Explainer}\label{subsec:ssye}
We introduce this embodiment in the main paper at the beginning of section \textbf{4.1} to better grasp the implicit minimization of $d_g$ and $d_a$ through the unique generation space and the symmetry constraint (see equation (\textbf{3}) in the main paper).\\

\noindent\textbf{Generators definition- }In the line of the one-to-one mapping used in \cite{shen_one--one_2020}, we consider this strict symmetrical version of SyCE (denoted SSyE). A unique generative model $g^{\star}$ defines the adversarial generator $\ga$. The same $g^{\star}$ is used for the two domains $\chi_0$ and $\chi_1$. 
\begin{equation}\label{eq:sym_adversary_supp}
	\ga(x) = g^{\star}(x), \, \, \, \forall x \in \chi
\end{equation}
Similar to SyCE, we impose a symmetry constraint on $g^*$ to define $\gs$.
\begin{equation}\label{eq:sym_similar_supp}
	\gs(x) = g^{2\star}(x), \, \, \, \forall x \in \chi
\end{equation}

\noindent\textbf{Visual explanation- }The visual explanation then reads 
\begin{equation}\label{eq:ExplanationDefg0g1SSyE_supp}
{\cal E}(x) = |g^{2\star} - g^{\star}(x)|
\end{equation} 

\noindent\textbf{Optimization problem- }In this formulation, the different terms of problem \eqref{eq:paperformulation_supp} are approximate in the same ways as in SyCE (except that there is a unique generator). The approximated problem reads

\begin{equation}\label{eq:SSyE_supp}
	g^{\star}  = \hspace{-0.9cm}
	\underset{
		\begin{array}{c}
			\scriptstyle g  \\
			\scriptstyle \text{s.t.} \\
			\scriptstyle g(\chi_0) \subset \chi_1, g(\chi_1) \subset \chi_0\\
			\scriptstyle g^2(\chi_0) \subset \chi_0, g^2(\chi_1) \subset \chi_1
		\end{array}
	}{\mathrm{argmin}} \hspace{-0.9cm}
	\begin{array}{c}
		\mathbb{E}_{x \in \chi} \left (||x -  g^2(x)|| \right ) 
	\end{array}
\end{equation}
\noindent\textbf{Framework- }In practice, the optimization framework is very close to the one for SyCE. We only remove the terms related to cycle consistency (i.e. $L_d^{cy}$ and $L_{f_c}^{cy}$) as there is a unique $g$ (See Figure \ref{fig:ssye_icml_supp}).

\subsection{CyCSAE: Cyclic Conditioned Stable Adversarial Explainer }\label{subsec:ssye}
\noindent\textbf{Generators definition- } we revisit and improve the "similar" / "adversarial" visual explanation method of \cite{Charachon2020CombiningSA}. 
We remove their explicit element-wise distances (adversarial-original) and (adversarial-similar). Instead, we add cyclic constraints to implicitly minimize them.

As in SyCE, we introduce condioned auxialiary generators such that:
\begin{equation}\label{eq:sym_adversary_CyCSAE_supp}
	\ga(x) = \left \{
	\begin{array}{l}
		g_{0}^{a\star}(x) \in \chi_1 \text{ if } x\in\chi_0 \\
	    g_{1}^{a\star}(x) \in \chi_0 \text{ if } x\in\chi_1 \\
	\end{array}\hspace{-0.1cm} 
	\right.
\end{equation}
And 
\begin{equation}\label{eq:sym_stable_CyCSAE_supp}
	\gs(x) = \left \{
	\begin{array}{l}
		g_{0}^{s\star}(x) \in \chi_0 \text{ if } x\in\chi_0 \\
	    g_{1}^{s\star}(x) \in \chi_1 \text{ if } x\in\chi_1 \\
	\end{array}\hspace{-0.1cm} 
	\right.
\end{equation}

\noindent\textbf{Visual explanation- }The visual explanation now reads :
\begin{equation}
	\Expl(x) = 
	\left \{\begin{array}{cc}
		|g_0^s(x) - g_0^a(x)|\, \text{ if }\,  x \in \chi_0 \\
		|g_1^s(x) - g_1^a(x)|\, \text{ if }\,  x \in \chi_1
	\end{array}
	\right \}
\end{equation}

\noindent\textbf{Optimization problem- }$d_a$ is enforced by a cycle consistency term as in CyCE. For $d_g$, we use a similar constraint $d_w$ as in \cite{Charachon2020CombiningSA} encouraging parameters of stable and adversarial generators ($g_i^s, \, \, g_i^a$ with $i \in \{0, 1\}$) to be close.
The approximated problem reads
\begin{equation}\label{eq:CyCSAE_supp}
\small
	(g_i^{s\star}, g_i^{a\star}) = \hspace{-0.9cm}
	\underset{
		\begin{array}{c}
			\scriptstyle g_i^{s}, g_i^{a} \\
			\scriptstyle i \in \{0,1\} \\
			\scriptstyle \text{s.t.} \\
			\scriptstyle g_0^s(\chi_0) \subset \chi_0, g_1^s(\chi_1) \subset \chi_1\\
			\scriptstyle g_0^a(\chi_0) \subset \chi_1, g_1^a(\chi_1) \subset \chi_0
		\end{array}
	}{\mathrm{argmin}} \hspace{-0.9cm}
	\sum_i\left[
	\begin{array}{l}
	 \mathbb{E}_{x \in \chi_i} \left (
		\begin{array}{l}
		   ||x - g_i^s(x)|| + \\
		   ||x - g_{|i-1|}^a(g_i^a(x))|| 
		\end{array} \right )   \\ + \\
		 d_w(g_i^s, g_i^a)
	\end{array}
	 \right]
\end{equation}
Where $g_{|i-1|}^a$ means $g_{1}^a$ if $i = 0$ (resp. $g_{0}^a$ if $i = 1$).
The unconstrained min-max approximation of \eqref{eq:CyCSAE_supp} remains quite similar to SyCE introduced in the paper.

\begin{figure}[h!]
    \begin{subfigure}[h]{\textwidth}
        \begin{subfigure}[h]{0.48\textwidth}
        \begin{center}
        \centerline{\includegraphics[width=\textwidth]{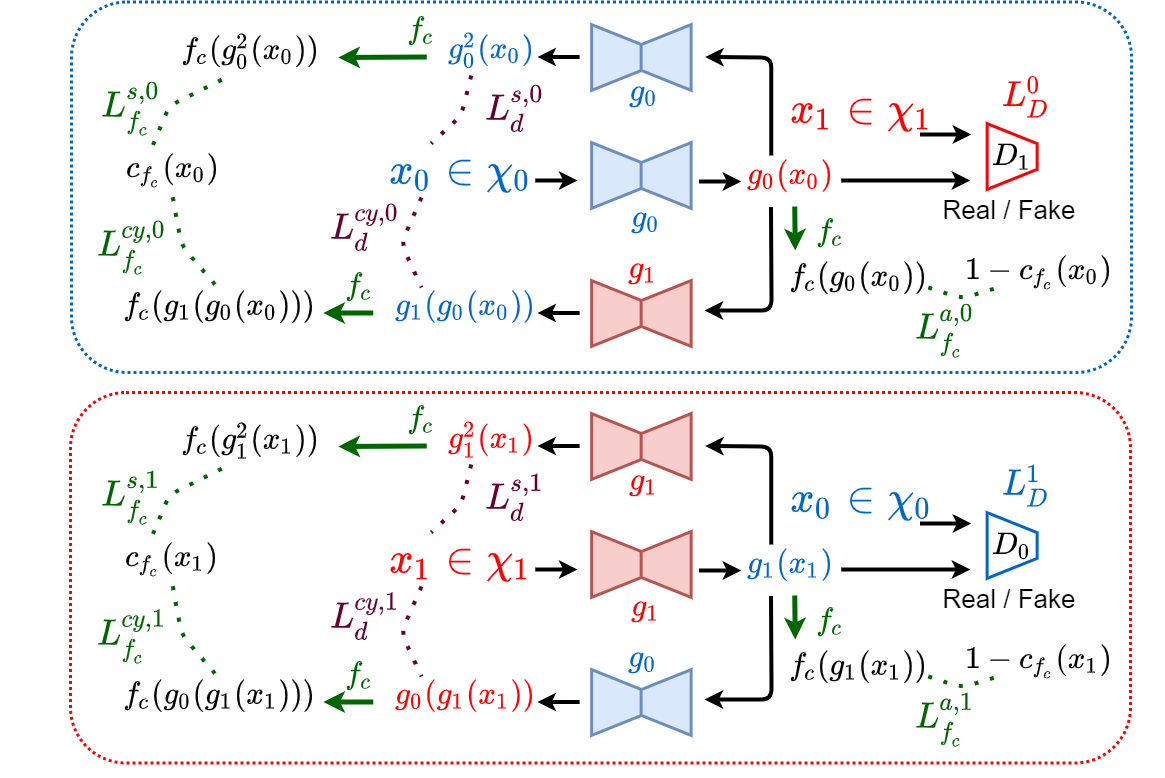}}
        \caption{\textbf{SyCE}}
        \label{fig:syce_icml_supp_supp}
        \end{center}
        \end{subfigure}
        \begin{subfigure}[h]{0.48\textwidth}
        \begin{center}
        \centerline{\includegraphics[width=\textwidth]{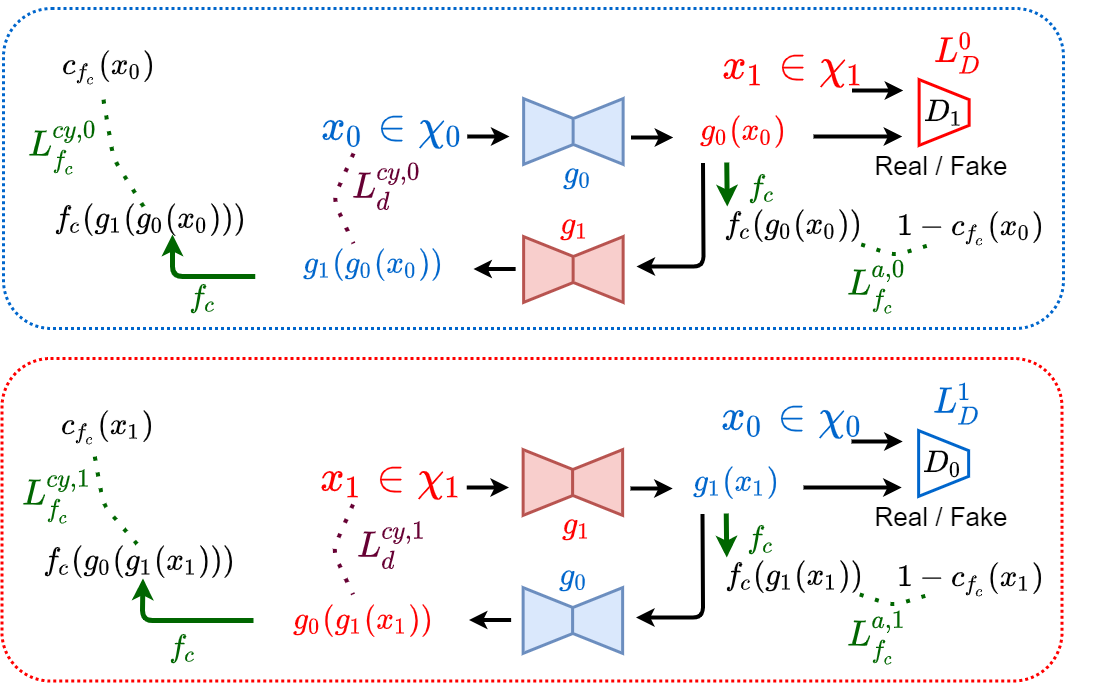}}
        \caption{\textbf{CyCE}}
        \label{fig:cyce_icml_supp}
        \end{center}
        \end{subfigure}
    \end{subfigure}
    \begin{subfigure}[h]{\textwidth}
        \begin{subfigure}[h]{0.48\textwidth}
        \begin{center}
        \centerline{\includegraphics[width=\textwidth]{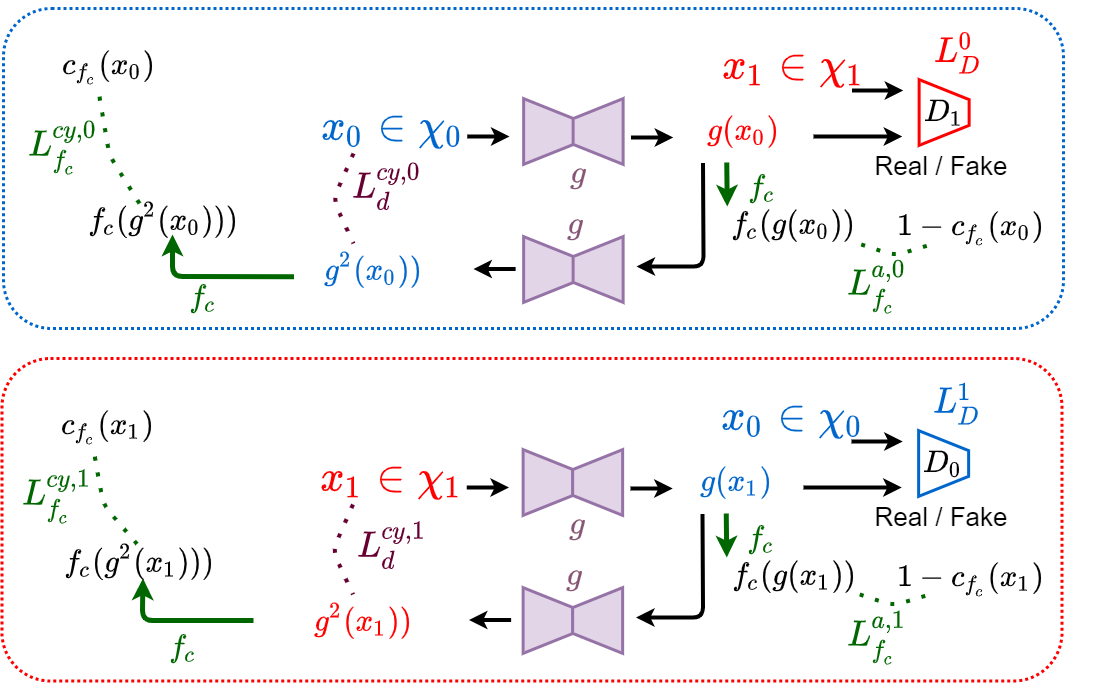}}
        \caption{\textbf{SSyE}}
        \label{fig:ssye_icml_supp}
        \end{center}
        \end{subfigure}
        \begin{subfigure}[h]{0.48\textwidth}
        \begin{center}
        \centerline{\includegraphics[width=\textwidth]{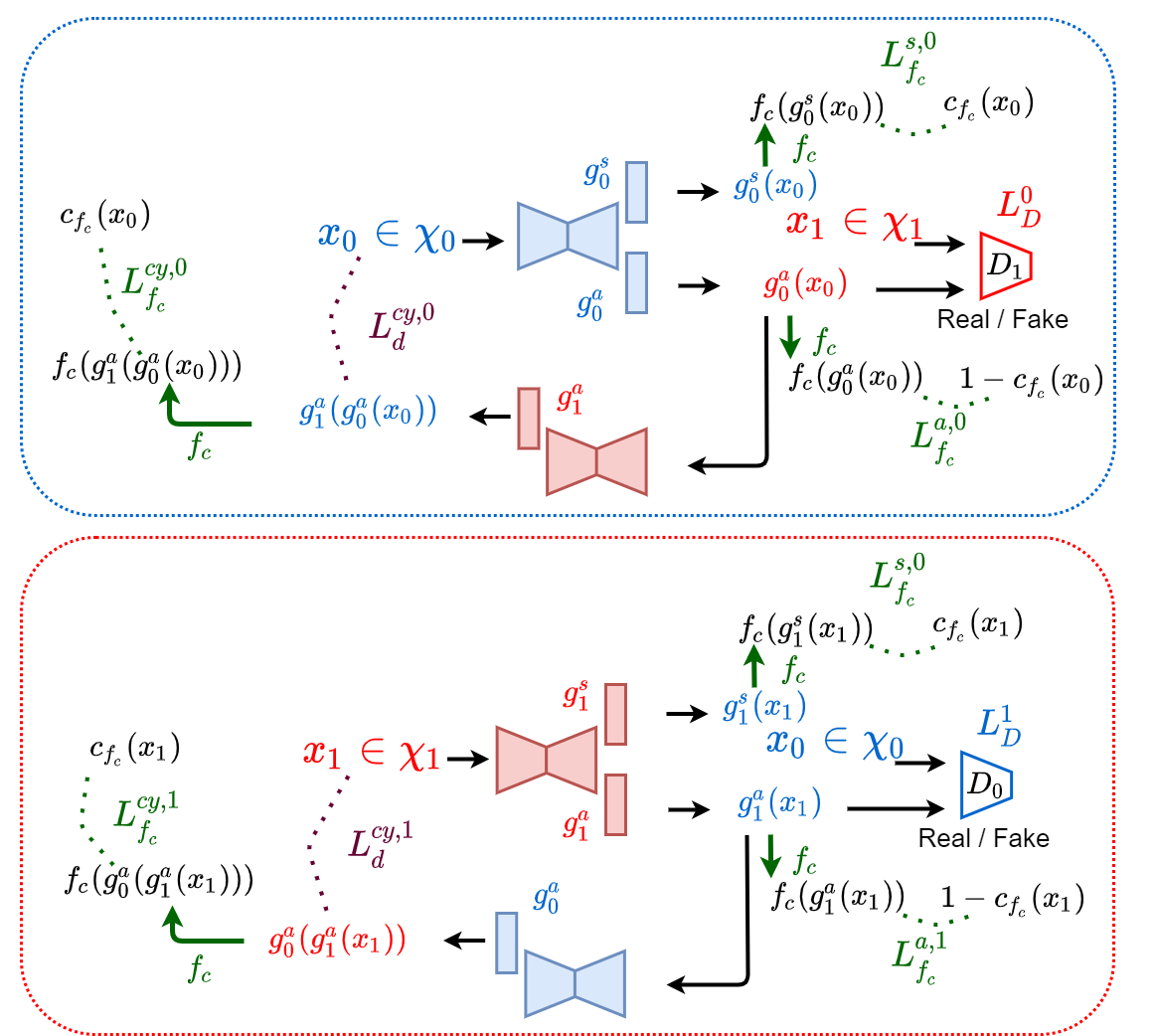}}
        \caption{\textbf{CyCSAE}}
        \label{fig:cycsae_icml_supp}
        \end{center}
        \end{subfigure}
    \end{subfigure}
    \caption{\textbf{Overview of the different optimization frameworks}. (a) \textbf{SyCE}. (b) \textbf{CyCE}. (c) \textbf{SSyE}. (d) \textbf{CyCSAE}. Top: training step of generators for an original image $x \in \chi_0$. Bottom: training step for $x \in \chi_1$.}
\end{figure}

\noindent\textbf{Framework- }In practice, $L_{Tot}^{CyCSAE} \approx L_{Tot}^{SyCE} + \lambda_w (d_w^0 + d_w^1)$. More precisely:
\begin{itemize}
    \item  \textit{\textbf{Adversarial Generation- }}Same $L_{f_c}^{a}$, $L_{D}$ are used with $g_{0}^a$ and $g_{1}^a$ instead of $g_{0}$ and $g_{1}$ (in SyCE)
    \item  \textit{\textbf{Cycle Consistency- }}Same $L_{d}^{cy}$, $L_{f_c}^{cy}$ are used with $g_{1}^a\circ g_{0}^a$ and $g_{0}^a\circ g_{1}^a$ instead of $g_{0}\circ g_{1}$ and $g_{1}\circ g_{0}$ (in SyCE).
    \item  \textit{\textbf{Stable Generation (instead of Symmetry in SyCE)- }}Same $L_{d}^{s}$, $L_{f_c}^{s}$ are used with $g_{0}^s$ and $g_{1}^s$ instead of $g_{0}^2$ and $g_{1}^2$ (in SyCE).
    \item \textit{\textbf{Generators Proximity Constraint- }}As in \cite{Charachon2020CombiningSA}, we use a $L_2$ constraint to enforce the proximity between parameters of $g_i^s$ ($w_i^s$) and $g_i^a$ ($w_i^a$), $i \in \{ 0, 1\}$
\end{itemize}
Figure \ref{fig:cycsae_icml_supp} illustrates the corresponding optimization.

\clearpage

% \end{document}
% \documentclass[../main.tex]{subfiles}
% \graphicspath{{\subfix{images/}}}

% \begin{document}

% \newcommand\f{{f_c}}
% \newcommand\thresh{\tau_{\f}}
% \newcommand\reals{\chi}
% \newcommand\E{{\chi_0}}
% \newcommand\F{{\chi_1}}
% \newcommand\cf{{c_{\f}}}
% \newcommand\gs{{g_{f_c}^{s\star}}}
% \newcommand\ga{{g_{f_c}^{a\star}}}
% \newcommand\gsopt{{g_{f_c}^{s}}}
% \newcommand\gaopt{{g_{f_c}^{a}}}
% \newcommand\Expl{{\cal E}}
% \newcommand\gz{{g_0^{\star}}}
% \newcommand\gu{{g_1^{\star}}}
%----------------------------------------------------------------------------------------
%	Multi-class Adaptation
%----------------------------------------------------------------------------------------
\section{Multi-class Adaptation}\label{sec:multiclass}
\subsection{General formulation}
In a multi-classification setting, the general formulation \eqref{eq:paperformulation} can be rewritten as:
\begin{equation}\label{eq:paperformulation}	
	(\gs, \ga) = \hspace{-0.7cm}
	\underset{
			\begin{array}{c}
				\scriptstyle \gsopt, \gaopt \\
				\scriptstyle \text{s.t. } 
				\left\{	
				\begin{array}{c}
					\scriptstyle \gsopt(\chi_i) \subset \mathbf{\chi_i} \\
					\scriptstyle \gaopt(\chi_i) \subset \mathbf{\bar{\chi_i}}
				\end{array}
				\right\}
			\end{array} }
		{\mathrm{argmin}} \hspace{-0.6cm}	
		\Big(
		\small
		d_g(\gsopt, \gaopt)+
		\mathbb{E}_x   \left (
		 +
		\begin{array}{c}
			d_s(x, \gsopt(x)) \\ 
			d_a(x, \gaopt(x)) 
		\end{array}
		\right )
		\Big)
\end{equation}
(i) $(d_g, d_a, d_s)$ are distances defined as in the general formulation \eqref{eq:paperformulation};\\
(ii) $\mathbf{\chi_i}$ is the subset of real images classified \textbf{as class i} by $\f$, they form a partition of $\reals$;\\
(iii) $\mathbf{\bar{\chi_i}}$ is the subset of real images classified \textbf{differently than class i} by $\f$. Combined with $\mathbf{\chi_i}$, they form a partition of $\reals$.

\subsection{Explanation objective}

\noindent\textbf{Explain One class against All others- }The idea is to explain one specific predicted class against all the others. In this case, the explanation problem is transformed into a binary case, where the objective is to inform why $\f$ predicted this class or not this class. Here, we can use similar embodiments as in the binary case. However, it might be expensive as we need to train one visual explainer for each class.\\

% The visual explanation for each class \textbf{i} would then read
% \begin{equation}\label{eq:ExplanationDefOneVSAll}
% 	\Expl_i(x) = |\gs_i(x) - \ga_i(x)|
% \end{equation}

\noindent\textbf{Explain Each class against Each others- }
Given an image $x$ and the classifier decision $\cf(x)$, a visual explanation should be produced to show for each different class why the classifier made this decision. For instance, in a classification problem with 10 classes, one classifier's decision requires 9 visual explanations (comparison with each different classes). Then, for an image $x$ predicted in \textbf{class i} by $\f$, the set of visual explanations reads
\begin{equation}\label{eq:ExplanationDefOneVSEach}
	\Expl_i(x) = \left \{|g_{f_c}^{s=i}(x) - g_{f_c}^{a=k}(x))| \right \}_{k \neq i, \, \, k \in \left [| 1, C  \right | ] }
\end{equation}
Where $C$ is the number of classes.

In this setting, it seems difficult to use embodiment such as SyCE directly, as we should train as many generators and discriminators as there are different classes. We should rather consider a unique couple of generator and discriminator,  but conditionned by $\f$ decision as conditional GANs \cite{Mirza2014ConditionalGA,Miyato2018cGANsWP}.

% \end{document}
% \documentclass[../main.tex]{subfiles}
% \graphicspath{{\subfix{images/}}}

% \begin{document}

%----------------------------------------------------------------------------------------
%	Datasets
%----------------------------------------------------------------------------------------
\section{Datasets}\label{sec:Datasets}
\subsection{Digits Identification - MNIST}\label{subsec:digits}
We design a binary classification task on the MNIST dataset \cite{lecun-mnisthandwrittendigit-2010} that consists in distinguishing digits "3" from digits "8". We extracted digits "3" and "8" from the original training dataset to create training and validation sets of respectively 9585 and 2397 samples. Then we similarly generate a test set of 1003 samples from the first 5,000 elements of the original test set. The original images of size 28 x 28 are normalized to [0, 1].

\subsection{Pneumonia detection - Chest X-Rays}\label{subsec:chestxrays}
We create a chest X-Rays dataset from the available RSNA Pneumonia Detection Challenge dataset which consists of X-Ray dicom exams extracted from the NIH CXR14 dataset \cite{DBLP:journals/corr/WangPLLBS17}. We only use the train images (26,684 exams) as we have access to class labels as well as expert annotations. The original dataset is composed of cases belonging to one of the three classes: \textbf{\textit{"Normal"}}, \textbf{\textit{"No Lung Opacity /Not Normal"}} and \textbf{\textit{"Lung Opacity"}}. For our task, we keep only the healthy (\textit{"Normal"}) and pathological (\textit{"Lung Opacity"}) exams that constitute a binary database of 14863 samples (8851 healthy / 6012 pathology). Pathological cases are provided with experts bounding box annotations around opacities. We then randomly split the training dataset into train (80 \%), validation (10 \%) and test (10 \%) sets. Original images are rescaled from 1024 x 1024 to 224 x 224 and then normalized to [0, 1].

\subsection{Brain Tumor localization - MRI}\label{subsec:brats}
Magnetic Resonance Imaging (MRI) is a medical application of nuclear magnetic resonance that generates 3D volumes of images. The localization of a tumor region on an MRI volume can be addressed as a binary classification problem. Indeed, we propose to classify each slice (2D image) along the axial axis as containing either one (or multiple) tumor region(s) [\textbf{class "1"}], or none [\textbf{class "0"}]. The brain MRI dataset comes from the  training set of the Medical Segmentation Decathlon Challenge \cite{Simpson2019ALA} which is composed of 484 exams. Each exam comes with 4 MRI modalities (FLAIR, T1w, T1gd, T2w) as well as 4 levels of segmentation annotations: background, edema, non-enhancing tumor and enhancing tumor. In this work, we only use the contrasted T1-weighted (T1gd) sequence. Ground truth mask annotation are computed by considering edema as background (\textbf{class "0"}), while non-enhancing and enhancing tumors are gathered in one single class: tumor (\textbf{class "1"}). First, we resample both the 3D T1gd volumes and the binary ground truth annotation masks from size 155 x 240 x 240 to 145 x 224 x 224. We extract the slices along the axial axis ($1^{st}$ axis), remove slices outside the brain (black images) and normalizing the images to [0, 1]. Then, we attribute a class label of "1" if a tumor region larger than 10 pixels (0.02 \% of the image size) exists on the corresponding annotation slice; and a label of "0" otherwise. The split consists of 363 training, 48 validation and 73 test patient exams which respectively correspond to 46900 - 6184 - 9424 slice images and the following class balancement: "0": 75 \% - "1": 25\%. 

% \end{document}
% \documentclass[../main.tex]{subfiles}
% \graphicspath{{\subfix{images/}}}

% \begin{document}

%----------------------------------------------------------------------------------------
%	Classifiers
%----------------------------------------------------------------------------------------

\section{Classifiers}\label{sec:Classifiers}
\subsection{Digits Identification - LeNet}\label{subsec:digits}
For this task we use a convolutional network based on LeNet \cite{lenet} with an adapted output for binary classification (sigmoid activation applied on an output layer of size 1). The model is trained with a binary cross entropy loss function for 50 epochs with a batch size of 128.

\subsection{Pneumonia detection - ResNet50 and DenseNet121}\label{subsec:chestxrays}
A ResNet50 \cite{DBLP:journals/corr/HeZR016} and a DenseNet121 \cite{Huang2017DenselyCC} are used here. We used pre-trained backbone layers from ImageNet \cite{imagenet_cvpr09} for the two models. The two models are trained with a binary cross entropy cost function for 50 epochs with batch size of 32.

\subsection{Brain Tumor localization - ResNet50 and DenseNet121}\label{subsec:brats}
We train similar ResNet50 and DenseNet121 with the same settings as the Chest X-Rays problem described above, except that they intend to minimize a weighted binary cross entropy. 

For all the problems, we use the Adam optimizer \cite{DBLP:journals/corr/KingmaB14} with an initial learning rate of 1e-4 which is divided by 3 each time a plateau is reached. Random geometric transformations such as zoom, translations, flips or rotations are introduced during training. The training of the classifier is run on 8 cpus Intel 52 Go RAM 1 V100, and last about 30 minutes for MNIST, 2 hours for Pneumonia detection and 3 hours for Brain tumor localization before convergence.

% \end{document}
% \documentclass[../main.tex]{subfiles}
% \graphicspath{{\subfix{images/}}}

% \begin{document}

%----------------------------------------------------------------------------------------
%	Implementation
%----------------------------------------------------------------------------------------
\section{Implementation}\label{subsec:Implementation}
\subsection{Ours: CyCE and SyCE}\label{subsec:ours}
\subsubsection*{Generators}\label{subsubsec:Generators}
For \textbf{CyCE} and \textbf{SyCE}, the generators $g_0$ and $g_1$ have the same structure and follow a UNet-like \cite{DBLP:journals/corr/RonnebergerFB15} architecture as suggested in \cite{pix2pix2017} for supervised image-to-image translation. The encoder path is composed of \textbf{3 blocks} of downsampling for the Brain tumor localization and the pneumonia detection problems. Only \textbf{2 blocks} are used for the digits identification. Each block consists of one residual block followed by a maxpooling layer. At each scale level of the decoded path, we concatenate the upsampled current layer with the corresponding skip connection layer from the encoder path. Several convolutions (2-3) are applied, then the resulting layer is upsampled to the next scale. Except for the final layer, convolution blocks are composed of batch normalization, ReLU activation and a convolution layer. Dropout (rate of 0.1 - 0.2 ) may be added after convolution blocks (better results achieved empirically). For the output layer, we either use a sigmoid activation or only clip the values in [0, 1] (so it can be passed to the trained classifier). Note that for the digits and brain tumor localization problems, better results are achieved with the clip.

\subsubsection*{Generators of other embodiments}\label{subsubsec:GeneratorsOthers}
(sec. \ref{sec:otherembodiment})

\noindent\textit{\textbf{SSyE- }}The unique generator $g$ is very similar to $g_0$ (and so $g_1$) in CyCE and SyCE. Experimentally, we observe that adding several convolution layers at the end of the UNet structure and before the output layer, slightly improve the optimization and localization resluts.

\noindent\textit{\textbf{CyCSAE- }}We use the shared architecture between $g_0^s$ and $g_0^a$ (resp. $g_1^s$ and $g_1^a$), as described in \cite{Charachon2020CombiningSA}. See sec. \ref{subsubsec:Baselines} for additional precision.

\subsubsection*{Discriminators}\label{subsubsec:Discriminators}
The discriminator models $D_0$ and $D_1$ also shared the same architecture which consists of \textbf{3} convolutional downsampling blocks (\textbf{2} for MNIST) followed by a dense linear layers that output a single logit vector (sigmoid cross entropy is used for the loss function). At each scale, the downsampling is achieved by a convolution layer with stride 2. We use LeakyReLU activation and no normalization layer. Note that using either ReLU or LeakyReLU with bacth normalization produce similar results.
For the specific "GAN" optimization, we use the common alternate optimization of corresponding pairs of generator and discriminator i.e. ($g_0$, $D_1$) and ($g_1$, $D_0$). In practice:
\begin{itemize}
    \item Rather than maximizing $L_D$, $g_0$ and $g_1$ are encouraged to minimize the following term $L_D^g$ with weighting parameters $\lambda_D$:
    \begin{equation}\label{eq:loss_Dg}
    \begin{array}{ll}
    \hspace{-0.25cm}L_{D}^g(x, g_0, g_1, D_0, D_1) \hspace{0.1cm}= 
         &\hspace{-0.25cm} \mathbb{E}_{x \in \chi_0} \hspace{0.1cm} L_{bce}(1, D_1(g_0(x)))  \hspace{0.25cm}+ \\
        & \hspace{-0.25cm} \mathbb{E}_{x \in \chi_1} \hspace{0.1cm}  \hspace{0.025cm} L_{bce}(1, D_0(g_1(x))) 
        \end{array}
    \end{equation}
    \item The discriminators $D_0$ and $D_1$ respectively minimize the term $L_D^d$ that adds a gradient penalty term $L_{gp}$ \cite{Gulrajani2017ImprovedTO}:
    \begin{equation}
        L_D^d = \lambda_D^d L_D + \lambda_{gp}^d L_{gp}
    \end{equation}
\end{itemize}

The models are trained on 8 cpus Intel 52 Go RAM 1 V100 for 80 epochs using Adam with an initial learning rate of 1e-4 for the generators and 2e-4 for the discriminators. We use a batch size of 64 for the MNIST dataset and 8 for the other problems. Random geometric transformations are applied with the same settings as for the classifier's training.  Optimization takes about 10 hours for CyCE and 15 hours SyCE (about the same for SSyE and CyCSAE).

\subsubsection*{Optimization Steps}\label{subsubsec:optimization_steps}

For both CyCE and SyCE:
\begin{itemize}
    \item We compute one optimization step for $g_0$ and $g_1$ given a batch of images $x$ in the source domain $\chi_0$, and $\chi_1$ the target domain. The transposed images $g_0(x)$ try to fool the discriminator $D_1$. Here $g_1$ is used for the cycle consistency.
    \item Symmetrically, we compute one optimization step for $g_1$ and $g_0$ for a batch of images $x \in \chi_1$.
    \item Then, we optimize the two discriminators $D_0$ and $D_1$ to respectively identify the batch of real images $x_0 \in \chi_0$ (resp. $x_1 \in \chi_1$) from generated images $g_1(x_1)$, $x_1 \in \chi_1$ (resp. $g_0(x_0)$, $x_0 \in \chi_0$).
\end{itemize}
The same optimization steps are used for SSyE and CyCSAE. In Figures \ref{fig:cyce_icml_supp}, \ref{fig:ssye_icml_supp} and \ref{fig:cycsae_icml_supp} we illustrate the global optimization framework of respectively CyCE, SSyE and CyCSAE (SyCE given in the paper).

\subsubsection*{Training Parameters}\label{subsubsec:training_params}

In tables \ref{tab:cycgan_params_supp} and \ref{tab:ours_params_supp} are given respectively the weighting parameters used for CyCE and SyCE. Those parameters are selected through empirical trials when achieving both the optimization objectives, an producing the best evaluation results. Note that for most of these parameters, variations within a certain interval produce similar results. 
\begin{table}[h!]
\caption{Training parameters for interpreters optimization}
\vskip 0.1in
\begin{subtable}[h]{\columnwidth}
\caption{CyCE}
\label{tab:cycgan_params_supp}
\begin{center}
\begin{small}
\begin{sc}
\begin{tabular}{lccccccc}
\toprule
Problem & Classifier& $\lambda_{d}^{cy}$ & $\lambda_{f_c}^{a}$ &  $\lambda_{f_c}^{cy}$ & $\lambda_{D}$ & $\lambda_{D}^d$ & $\lambda_{gp}^d$\\
\midrule
\midrule
Digits&LeNet&10.0&0.2&0.005&0.25&1.0&1.0 \\
\midrule
\multirow{2}{*}{Pneumonia detect.}&ResNet50 &10.0&0.05&0.01&0.025&1.0&1.0 \\
&DenseNet121 &10.0&0.05&0.01&0.05&1.0&1.0 \\
\midrule
\multirow{2}{*}{Brain tumor loc}&ResNet50&20.0&0.1&0.01&0.05&1.0&1.0 \\
&DenseNet121 &20.0&0.2&0.01&0.05&1.0&1.0 \\
\bottomrule
\end{tabular}
\end{sc}
\end{small}
\end{center}
\end{subtable}
\vskip 0.1in
\begin{subtable}[h]{\columnwidth}
\caption{SyCE}
\label{tab:ours_params_supp}
\begin{center}
\begin{small}
\begin{sc}
\begin{tabular}{lccccccccc}
\toprule
Problem & Classifier&$\lambda_{d}^{s}$ & $\lambda_{d}^{cy}$ & $\lambda_{f_c}^{a}$ & $\lambda_{f_c}^{s}$ & $\lambda_{f_c}^{cy}$ & $\lambda_{D}$ & $\lambda_{D}^d$ & $\lambda_{gp}^d$\\
\midrule
\midrule
Digits&LeNet&10.0&2.0&0.2&0.01&0.005&0.25&1.0&1.0 \\
\midrule
\multirow{2}{*}{Pneumonia detect.}&ResNet50 &10.0&2.0&0.05&0.01&0.001&0.05&1.0&1.0 \\
&DenseNet121 &10.0&2.0&0.05&0.01&0.01&0.05&1.0&1.0 \\
\midrule
\multirow{2}{*}{Brain tumor loc}&ResNet50&20.0&1.0&0.2&0.05&0.01&0.025&1.0&1.0 \\
&DenseNet121 &20.0&1.0&0.2&0.05&0.001&0.025&1.0&1.0 \\
\bottomrule
\end{tabular}
\end{sc}
\end{small}
\end{center}
\vskip -0.1in
\end{subtable}
\end{table}

% \begin{table}[ht]
% \caption{Training paramters used for our interpreter's optimization}
% \label{tab:ours_params}
% \vskip 0.1in
% \begin{adjustbox}{width=\columnwidth,center}
% \begin{small}
% \begin{sc}
% \begin{tabular}{lccccccccc}
% \toprule
% Problem & Classifier&$\lambda_{d}^{s}$ & $\lambda_{d}^{cy}$ & $\lambda_{f_c}^{a}$ & $\lambda_{f_c}^{s}$ & $\lambda_{f_c}^{cy}$ & $\lambda_{D}$ & $\lambda_{D}^d$ & $\lambda_{gp}^d$\\
% \midrule
% \midrule
% Digits&LeNet&10.0&2.0&0.2&0.01&0.005&0.25&1.0&1.0 \\
% \midrule
% \multirow{2}{*}{Pneumonia detect.}&ResNet50 &10.0&2.0&0.05&0.01&0.001&0.05&1.0&1.0 \\
% &DenseNet121 &10.0&2.0&0.05&0.01&0.01&0.05&1.0&1.0 \\
% \midrule
% \multirow{2}{*}{Brain tumor loc}&ResNet50&20.0&1.0&0.2&0.05&0.01&0.025&1.0&1.0 \\
% &DenseNet121 &20.0&1.0&0.2&0.05&0.001&0.025&1.0&1.0 \\
% \bottomrule
% \end{tabular}
% \end{sc}
% \end{small}
% \end{adjustbox}
% \vskip -0.1in
% \end{table}

\subsection{Baselines Methods}\label{subsubsec:Baselines}
\textbf{BBMP} \cite{fong_interpretable_2017}- The best explanation maps are obtained when we start with a mask of size 56 x 56, and then filter the upsampled mask (gaussian with $\sigma = 3$) such as in \cite{fong_interpretable_2017}. We generate the explanation mask after 150 iterations. We also use a total variation regularization. The gaussian blur perturbation ($\sigma = 5$) produces the better results. \\
\textbf{MGen} \cite{dabkowski_real_2017}- A ResNet-50 backbone pre-trained on ImageNet (or on the task) is used for the encoder part of the model as in \cite{dabkowski_real_2017}. As suggested in \cite{Charachon2020CombiningSA}, we adapt the architecture of MGen to a binary classification problem, and remove the class embedding input that does not improve the mask generation. For the training, we follow the directions proposed in \cite{dabkowski_real_2017}. More specifically, we alternate between gaussian blur perturbation and a mix of constant and random noise. The generator model produces masks of size 112x112 (14x14 for MNIST) that are upsampled to 224x224 (28x28 for MNIST). As in BBMP, an additional gaussian filtering helps to remove some artefacts.\\
\textbf{SAGen} \cite{Charachon2020CombiningSA}- follows the single generator version described in \cite{Charachon2020CombiningSA}. In practice, the architecture consists of a common UNet-like architecture (similar architecture than our generators) and two separated final convolutional blocks: one for the stable image generation and the other for the adversarial image generation.

% \end{document}
% \documentclass[../main.tex]{subfiles}
% \graphicspath{{\subfix{images/}}}

% \begin{document}
%----------------------------------------------------------------------------------------
%	Results
%----------------------------------------------------------------------------------------

\section{Results}\label{sec:results}

\subsection{Evaluation of domain translation}\label{subsec:Domaintranslation}

\textbf{Comparison with other embodiments- }We compare domain translation achieved by the different generation methods, including embodiments SSyE and CyCSAE.

\begin{figure*}[ht]
\begin{subfigure}[b]{\textwidth}
	\begin{center}
		\centerline{\includegraphics[width=\textwidth]{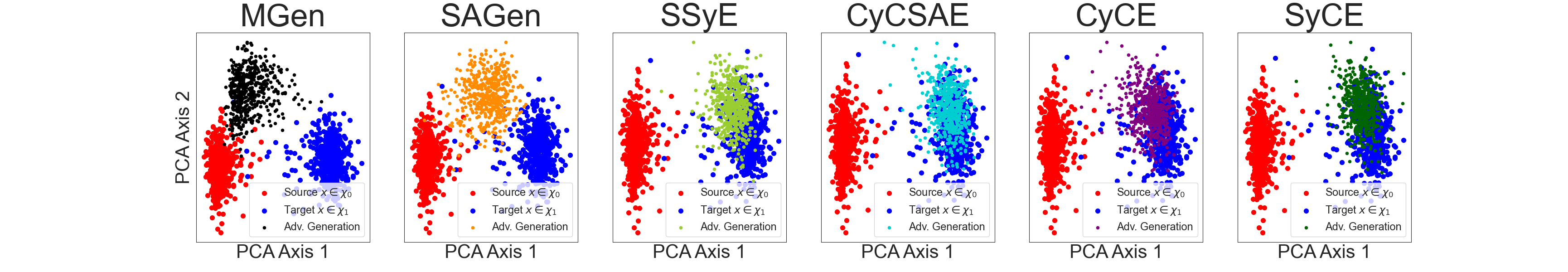}}
		\vskip -0.1in
		\caption{\textbf{\textcolor{red}{$\chi_0$} $\longrightarrow$ \textcolor{blue}{$\chi_1$} - Adversarial Generation}}
		\label{fig:pca_mnist_01_adv_supp}
	\end{center}
\end{subfigure}
\vskip -0.05in
\begin{subfigure}{\textwidth}
	\begin{center}
		\centerline{\includegraphics[width=\textwidth]{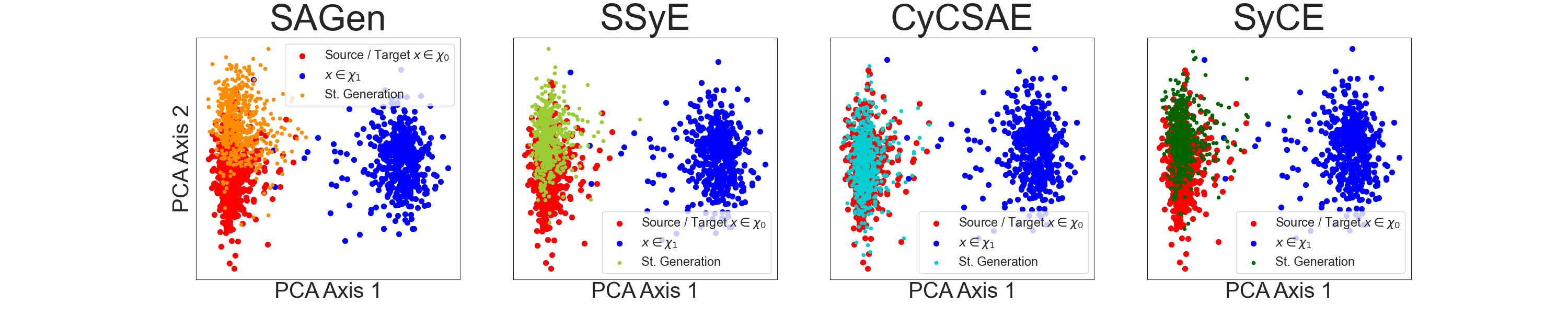}}
		\vskip -0.1in
		\caption{\textbf{\textcolor{red}{$\chi_0$} $\longrightarrow$ \textcolor{red}{$\chi_0$} - Stable Generation}}
		\label{fig:pca_mnist_01_sim_supp}
	\end{center}
\end{subfigure}
\vskip 0.2in
\begin{subfigure}[b]{\textwidth}
	\begin{center}
		\centerline{\includegraphics[width=\textwidth]{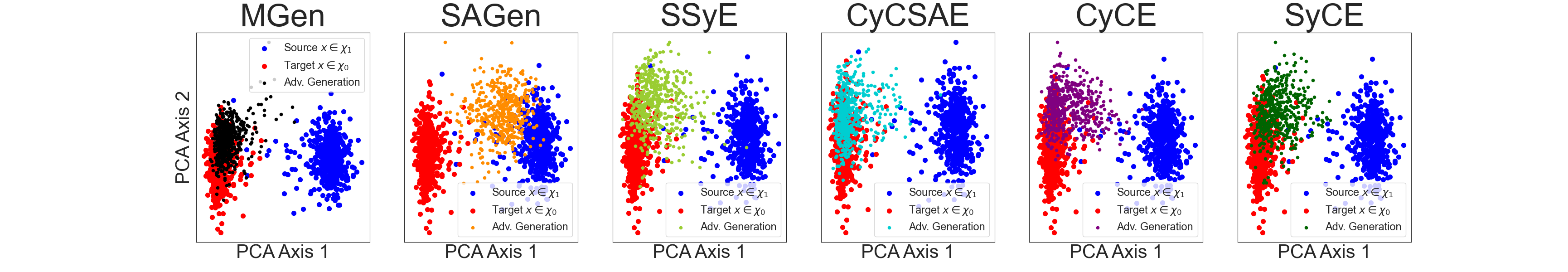}}
		\vskip -0.1in
		\caption{\textbf{\textcolor{blue}{$\chi_1$} $\longrightarrow$ \textcolor{red}{$\chi_0$} - Adversarial Generation}}
		\label{fig:pca_mnist_10_adv_supp}
	\end{center}
\end{subfigure}
\vskip -0.05in
\begin{subfigure}{\textwidth}
	\begin{center}
		\centerline{\includegraphics[width=\textwidth]{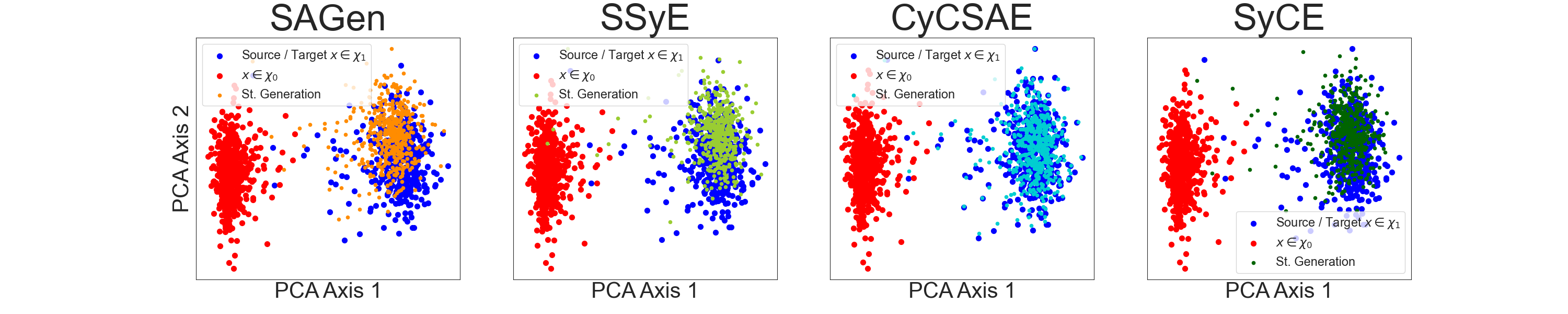}}
		\vskip -0.1in
		\caption{\textbf{\textcolor{blue}{$\chi_1$} $\longrightarrow$ \textcolor{blue}{$\chi_1$} - Stable Generation}}
		\label{fig:pca_mnist_10_sim_supp}
	\end{center}
\end{subfigure}
% \vskip -0.1in
	\caption{First 2 axes of the PCA applied on the embedded vector $\mu$ of the VAE for all images (real and generated) of the test set for \textbf{MNIST digits classification}. From top to bottom: Adversarial generation from Source (original) domain $\chi_0$ ("3") to target domain $\chi_1$ ("8"); Stable generation in $\chi_0$; Adversarial generation from $\chi_1$ ("8") to $\chi_0$ ("3");  Stable generation in $\chi_1$}
\vskip -0.1in
\end{figure*}

Figures \ref{fig:pca_mnist_01_adv_supp}, \ref{fig:pca_mnist_01_sim_supp}, \ref{fig:pca_mnist_10_adv_supp} and \ref{fig:pca_mnist_10_sim_supp} shows the 2-dimensional PCA representation of all generated and real images for the digits problem. 
\begin{itemize}
\item All approximations of \textbf{our general formulation} achieve adversarial generations if both space translations.
% to produce adversarial images in the opposite image distribution in the two directions "3" to "8" and "8" to "3". 
\item SAGen \cite{Charachon2020CombiningSA} (orange dots) fails to produce adversaries in the opposite distribution. Instead, they remain close to original images. 
\item MGen (black) achieves to generate "3" from "8", but completely fails in the other direction.
\end{itemize}

\begin{figure*}[ht]
\begin{subfigure}[b]{\textwidth}
	\begin{center}
		\centerline{\includegraphics[width=\textwidth]{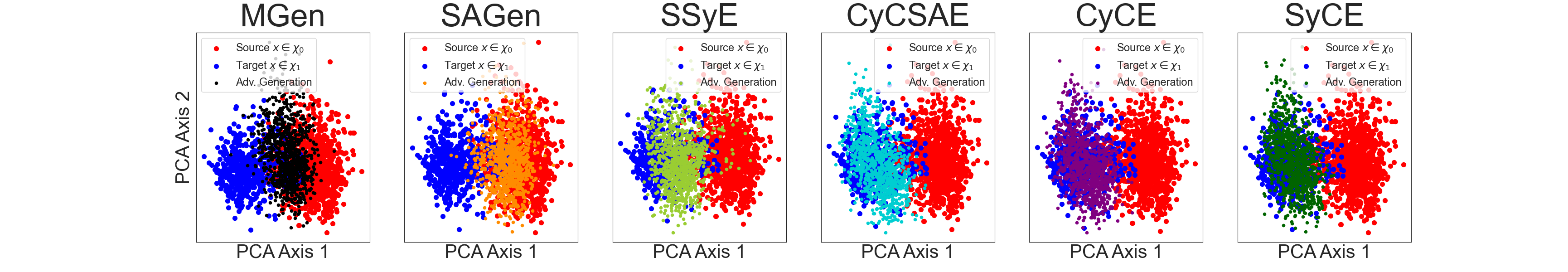}}
		\vskip -0.1in
		\caption{\textbf{\textcolor{red}{$\chi_0$} $\longrightarrow$ \textcolor{blue}{$\chi_1$} - Adversarial Generation}}
		\label{fig:pca_xray_01_adv_supp}
	\end{center}
\end{subfigure}
\vskip -0.05in
\begin{subfigure}{\textwidth}
	\begin{center}
		\centerline{\includegraphics[width=\textwidth]{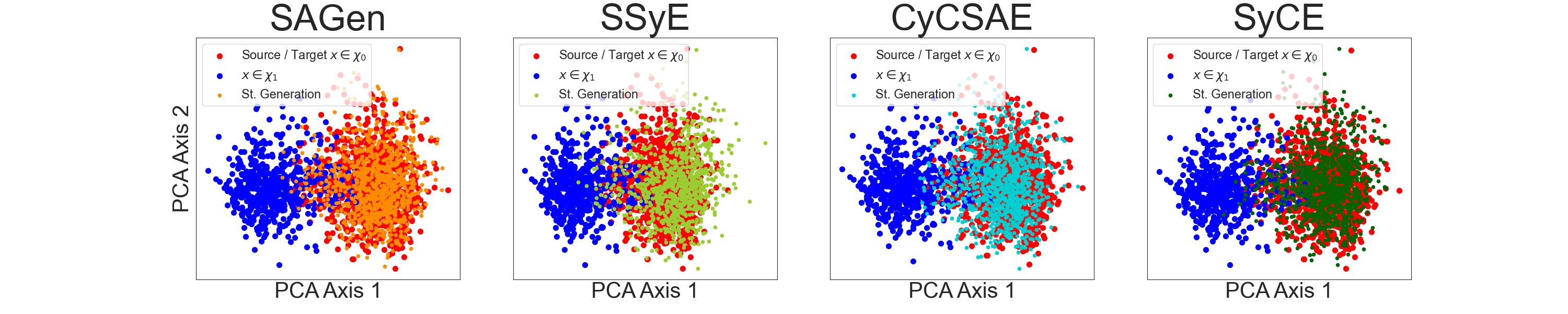}}
		\vskip -0.1in
		\caption{\textbf{\textcolor{red}{$\chi_0$} $\longrightarrow$ \textcolor{red}{$\chi_0$} - Stable Generation}}
		\label{fig:pca_xray_01_sim_supp}
	\end{center}
\end{subfigure}
\vskip 0.2in
\begin{subfigure}[b]{\textwidth}
	\begin{center}
		\centerline{\includegraphics[width=\textwidth]{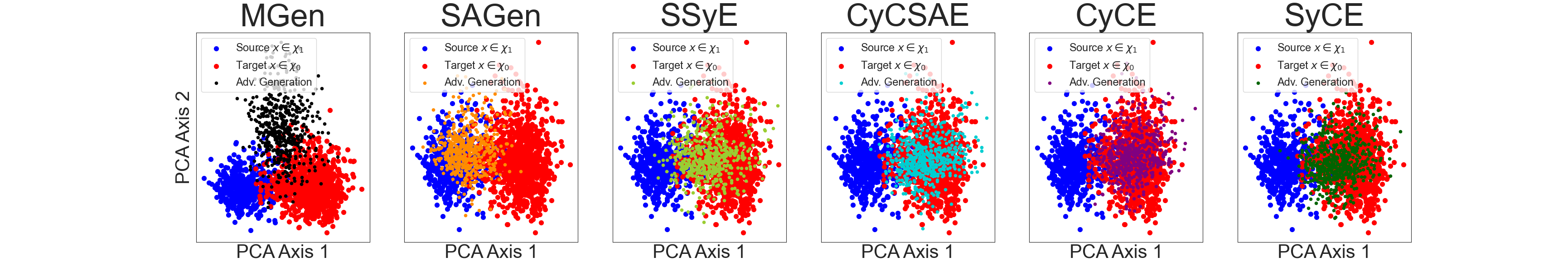}}
		\vskip -0.1in
		\caption{\textbf{\textcolor{blue}{$\chi_1$} $\longrightarrow$ \textcolor{red}{$\chi_0$} - Adversarial Generation}}
		\label{fig:pca_xray_10_adv_supp}
	\end{center}
\end{subfigure}
\vskip -0.05in
\begin{subfigure}{\textwidth}
	\begin{center}
		\centerline{\includegraphics[width=\textwidth]{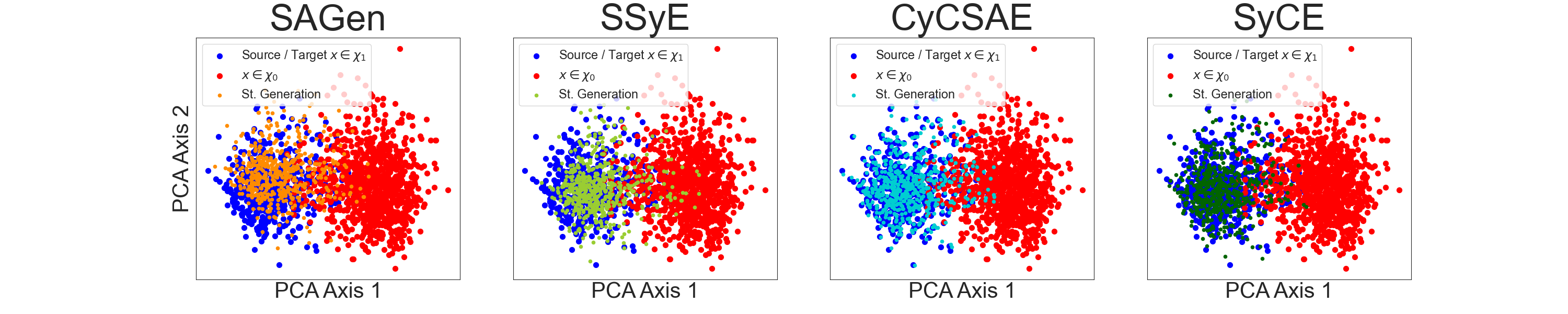}}
		\vskip -0.1in
		\caption{\textbf{\textcolor{blue}{$\chi_1$} $\longrightarrow$ \textcolor{blue}{$\chi_1$} - Stable Generation}}
		\label{fig:pca_xray_10_sim_supp}
	\end{center}
\end{subfigure}
% \vskip -0.1in
	\caption{First 2 axes of the PCA applied on the embedded vector $\mu$ of the VAE for all images (real and generated) of the test set for \textbf{Pneumonia detection}. From top to bottom: Adversarial generation from Source (original) domain $\chi_0$ ("Healthy") to target domain $\chi_1$ ("Pathological"); Stable generation in $\chi_0$; Adversarial generation from $\chi_1$ ("Pathological") to $\chi_0$ ("Healthy");  Stable generation in $\chi_1$}
\vskip -0.1in
\end{figure*}

Figures \ref{fig:pca_xray_01_adv_supp}, \ref{fig:pca_xray_01_sim_supp}, \ref{fig:pca_xray_10_adv_supp} and \ref{fig:pca_xray_10_sim_supp} present the same representation for the X-Rays pneumonia detection. 
\begin{itemize}
\item SyCE (green dots), CyCE (purple dots) and CyCSAE (cyan dots) generate adversarial images close to the opposite distribution
\item SSyE (lime dots) is slighlty less efficient in the adversarial domain translation.
\end{itemize}

Table \ref{tab:classif-performances_supp} describes the classification performances of stable and adversarial generation compared to original decision $c_{f_c}(x)$.
\begin{itemize}
    \item All guided methods succeed in generating adversarial images for the classifier $f_c$, while 
    \item CyCE w/o $L_{f_c}^{a, cy}$ (similar to \cite{Narayanaswamy2020ScientificDB}) produces much poorer classification results when generating adversarial examples for $f_c$.
    \item Stable generations (for methods that produce some) retain the classification score quite well.
\end{itemize}

These observations are supported by the results shown in Tables \ref{tab:chi02chi1_others_supp} and \ref{tab:chi12chi0_others_supp}. 
Accross all experiments
\begin{itemize}
    \item CyCSAE outperforms other methods for MNIST
    \item CyCE is the best performer on both chest X-Rays and brain MRI problems
    \item SSyE produces poorer domain translation results than other embodiments
\end{itemize}
\newpage

\begin{table}[ht]
\caption{\textbf{Classification performances}. Accuracies of respectively LeNet on Digits and ResNet on Pneumonia detection and Brain tumor localization: $acc_s$ and $acc_a$ on respectively stable and adversarial generated images.
The accuracy $acc_s$ is computed between the original model's decision $c_{f_c}(x)$ and the decision on the stable image ($c_{f_c}(g_s(x))$). $acc_a$ is computed between the original model's decision and the decision on the adversarial image.
}
\label{tab:classif-performances_supp}
% \begin{adjustbox}{width=\columnwidth,center}
\begin{center}
\begin{small}
\begin{sc}
\begin{tabular}{lcccccc}
\toprule
Method & \multicolumn{2}{c}{Digits} & \multicolumn{2}{c}{Pneumonia} & \multicolumn{2}{c}{Tumor Loc.} \\
& $acc_s \uparrow$& $acc_a \downarrow$ & $acc_s\uparrow$& $acc_a \downarrow$ & $acc_s\uparrow$& $acc_a \downarrow$ \\
\midrule
Mgen&-&0.176&-&0.075&-&0.156 \\
SAGen&0.991&0.032&0.981&0.103&0.957&0.243    \\
CyCE w/o $L_{f_c}^{a, cy}$ &-&0.954&-&0.739&-&0.966\\
CyCE &-&0.070&-&0.040&-&\textbf{0.090}\\
SyCE&\textbf{0.993}&\textbf{0.015}&\textbf{0.988}&\textbf{0.028}& \textbf{0.966} &0.096   \\
\bottomrule
\end{tabular}
\end{sc}
\end{small}
\end{center}
\end{table}

\begin{table}[h!]
\caption{\textbf{Domain Translation results} - LeNet for MNIST digits and ResNet 50 for medical imaging problems.}
\begin{subtable}[h]{0.48\textwidth}
\caption{$\chi_0 \longrightarrow \chi_1$}
\label{tab:chi02chi1_others_supp}
\begin{adjustbox}{width=\columnwidth,center}
\begin{small}
\begin{sc}
\begin{tabular}{lccccccc}
\toprule
Method && \multicolumn{2}{c}{Digits} & \multicolumn{2}{c}{Pneumonia} & \multicolumn{2}{c}{Tumor Loc.} \\
& &$FD_{\mu} \downarrow$& $JS_{\mu} \downarrow$&$FD_{\mu} \downarrow$& $JS_{\mu} \downarrow$&$FD_{\mu} \downarrow$& $JS_{\mu} \downarrow$\\
\midrule
\midrule
Mgen& A&142.71& 0.99&58.76&0.79&65.29&0.46\\
\midrule
\multirow{2}{*}{SAGen}&\textit{St}&\textit{(1.67)}& \textit{(0.68)}&\textit{(0.22)}& \textit{(0.09)}&\textit{(1.54)}& \textit{(0.14)}\\
&A&55.93&  0.96 & 89.81&0.85&242.22&0.72\\
\midrule
\multirow{2}{*}{SSyE}&\textit{St}&\textit{(0.39)}&\textit{(0.49)}&\textit{(3.00)}&\textit{(0.25)}&\textit{(0.01)}& \textit{(0.08)}\\
&A&4.88& 0.62&9.05&0.44 & 133.49&0.55\\
\midrule
\multirow{2}{*}{CyCSAE}&\textit{St}&\textit{(0.01)}&\textit{(0.07)}&\textit{(0.42)}&\textit{(0.10)}& \textit{(0.04)}& \textit{(0.03)}\\
&A&\textbf{1.22}&\textbf{0.42}&2.16& 0.36&50.77 &0.47\\
\midrule
CyCE&A&4.50&0.54&\textbf{1.92}&\textbf{0.34}&\textbf{2.66}&\textbf{0.29}\\
\midrule
\multirow{2}{*}{SyCE}&\textit{St}&\textit{(0.30)}&\textit{(0.40)}&\textit{(0.04)}&\textit{(0.08)}& \textit{(0.13)}& \textit{(0.07)}\\
&A&2.84& 0.56&2.12& 0.35&45.88&0.45\\
\bottomrule
\end{tabular}
\end{sc}
\end{small}
\end{adjustbox}
\end{subtable}
\begin{subtable}[h]{0.48\textwidth}
\caption{$\chi_1 \longrightarrow \chi_0$}
\label{tab:chi12chi0_others_supp}
\begin{adjustbox}{width=\columnwidth,center}
\begin{small}
\begin{sc}
\begin{tabular}{lccccccc}
\toprule
Method && \multicolumn{2}{c}{Digits} & \multicolumn{2}{c}{Pneumonia} & \multicolumn{2}{c}{Tumor Loc.} \\
& &$FD_{\mu} \downarrow$& $JS_{\mu} \downarrow$&$FD_{\mu} \downarrow$& $JS_{\mu} \downarrow$&$FD_{\mu} \downarrow$& $JS_{\mu} \downarrow$ \\
\midrule
\midrule
Mgen& A&3.10& 0.61&38.06&0.84&94.70&0.56\\
\midrule
\multirow{2}{*}{SAGen}&\textit{St}&\textit{(0.11)}& \textit{(0.39)}&\textit{(0.84)}& \textit{(0.27)}&\textit{(10.97)}&\textit{(0.11)}\\
&A&12.04&  0.99 & 95.17&0.85&312.64&0.77\\
\midrule
\multirow{2}{*}{SSyE}&\textit{St}&\textit{(0.34)}&\textit{(0.36)}&\textit{(1.04)}&\textit{(0.20)}&\textit{(0.14)}&\textit{(0.04)}\\
&A&13.00& 0.79& 14.75&0.45 & 281.68&0.74\\
\midrule
\multirow{2}{*}{CyCSAE}&\textit{St}&\textit{(0.01)}&\textit{(0.06)}&\textit{(0.06)}&\textit{(0.04)}&\textit{(0.004)}&\textit{(0.02)}\\
&A&\textbf{3.09}& \textbf{0.50}& 3.33&0.31 & 69.70&0.47\\
\midrule
CyCE&A&9.37& 0.74&\textbf{1.56}&\textbf{0.28}&\textbf{39.24}&\textbf{0.41}\\
\midrule
\multirow{2}{*}{SyCE}&\textit{St}&\textit{(0.30)}&\textit{(0.27)}&\textit{(0.36)}&\textit{(0.11)}&\textit{(0.05)}&\textit{(0.05)}\\
&A&8.61& 0.72&9.71& 0.41&51.51&0.44\\
\bottomrule
\end{tabular}
\end{sc}
\end{small}
\end{adjustbox}
\end{subtable}
\end{table}

\clearpage

\clearpage
\subsection{Relevant regions found by the visual explanation}\label{subsec:FA}
\textbf{Metrics and thresholds choices- }We consider two metrics to evaluate the localization performance of visual explanation techniques: The intersection over union ($IoU$) and the normalized cross-correlation ($NCC$).
As mentioned in the paper, the $IoU$ is computed between binary ground truth annotation
(e.g. filled bounding boxes for pneumonia detection, and segmentation masks for tumor localization) and a thresholded binary explanation mask. 

The choice of the threshold depends on the representative size of the annotations (w.r.t the image size) on the dataset.
Tables \ref{tab:stats_xray_supp} and \ref{tab:stats_mri_supp} show the main statistics of the size of expert annotations on the training set. We give the results as ratios of the size of the image. \\
For \textbf{Pneumonia expert annotations}, bounding box annotations represent 8.8 \% (median) of the image size whatever the number of pathological regions annotated; 4.4 \% when there is a unique bounding box and about 14.6\% when there are at least 2 different annotations. In addition, bounding box are weaker annotations than segmentation mask as they also contain regions that are not included in the opacity (pneumonia signature). 
\begin{table}[h!]
\caption{Annotations statistics - Some basic statistics about the size of the ground truth annotations on the \textbf{training set}.The figures are given as ratios (\%) to the size of the image.}
\begin{subtable}[h]{\textwidth}
\caption{Pneumonia Detection - Chest X-rays}
\label{tab:stats_xray_supp}
\begin{center}
\begin{small}
\begin{sc}
\begin{tabular}{lcccccccc}
\toprule
Annotations&Nb&Mean&Median&Std&Min&Max&$25^{th}$ Perc.&$75^{th}$ Perc.\\
\midrule
\multirow{3}{*}{Bounding Box}&All&11.7&8.8&9.4&0.3&60.2&4.5&16.5\\
&$1$&5.4&4.4&3.9&0.3&35.3&2.6&7.1\\
&$> 1$&16.6&14.6&9.5&1.4&60.2&9.1&22.7\\
\bottomrule
\end{tabular}
\end{sc}
\end{small}
\end{center}
\end{subtable}
\begin{subtable}[h]{\textwidth}
\vskip 0.2in
\caption{Brain Tumor Localization - MRI}
\label{tab:stats_mri_supp}
\begin{center}
\begin{small}
\begin{sc}
\begin{tabular}{lcccccccc}
\toprule
Annotations&Nb&Mean&Median&Std&Min&Max&$25^{th}$ Perc.&$75^{th}$ Perc.\\
\midrule
Segmentation&all ($=1$)&1.5&1.2&1.2&0.04&6.3&0.5&2.2\\
\bottomrule
\end{tabular}
\end{sc}
\end{small}
\end{center}
\end{subtable}
\end{table}

To capture the variability of the size of the annotations (single and multiple), we choose the following thresholds: the 85th, 90th, 95th and 98th percentiles. These choices match with the bounding box statistics and act for annotations that cover from 2 to 15\% of the size of the image. E.g. For a binary explanation map thresholded at the 95th percentile, the corresponding $IoU$ is denoted $IoU_{95}$.\\
Similarly for the \textbf{brain mri problem}, we set thresholds at the 97th, 98th and 99th percentiles for explanation maps. Table \ref{tab:stats_mri_supp} shows that segmentation masks of the tumors have a much smaller variability i.e.  1.5 $\pm$ 1.2 \%.

\textbf{Localization results- }Tables \ref{tab:loc_results_xray_resnet_supp} and \ref{tab:loc_results_xray_densenet_supp} display the localization results on the X-Ray pneumonia detection problem for the two metrics and for the two different classifiers (ResNet-50 in tab. \ref{tab:loc_results_xray_resnet_supp} and DenseNet-121 in tab. \ref{tab:loc_results_xray_densenet_supp}). The same results are given for brain MRI tumor localization in Tables \ref{tab:loc_results_mri_resnet_supp} and \ref{tab:loc_results_mri_densenet_supp}.
\begin{table}[h!]
\caption{\textbf{Localization results for X-Rays Pneumonia detection}. Comparison with state-of-the-art techniques and other embodiments (SSyE and CyCSAE).}
\begin{subtable}[h]{0.48\textwidth}
\caption{Results for ResNet-50}
\label{tab:loc_results_xray_resnet_supp}
\begin{adjustbox}{width=\columnwidth,center}
\begin{small}
\begin{sc}
\begin{tabular}{lcccccc}
\toprule
Method & $IoU_{85}$& $IoU_{90}$& $IoU_{95}$ & $IoU_{98}$ & NCC \\
\midrule
\midrule
Gradient&0.199&0.187&0.152&0.097&0.312\\
IG&0.171& 0.170&0.136&0.086&0.254\\
GradCAM&0.224& 0.195&0.138&0.070&0.325\\
BBMP&0.226&0.204&0.154&0.087& 0.348\\
Mgen&0.219&0.208&0.169&0.103& 0.340\\
SAGen & 0.250&0.232&0.173&0.097&0.325 \\
\midrule
SSyE w/o St.&0.222&0.211&0.168&0.095&0.309\\
SSyE&0.240&0.230&0.183&0.107&0.363 \\
\midrule
CyCSAE w/o St.&0.290&0.286&0.233&0.142&0.453 \\
CyCSAE&0.294&0.291&0.238&0.147&0.461 \\
\midrule
CyCE&0.219&0.221&0.191&0.116&0.337 \\
\midrule
SyCE w/o $L^{cy}$&0.277&0.260&0.205&0.118&0.443\\
SyCE w/o St.&0.313& 0.294&0.238&0.142&0.498\\
SyCE &   \textbf{0.316}&\textbf{0.299}&\textbf{0.244}&\textbf{0.151}&  \textbf{0.506}\\
\bottomrule
\end{tabular}
\end{sc}
\end{small}
\end{adjustbox}
\end{subtable}
\begin{subtable}[h]{0.48\textwidth}
\caption{Results for DenseNet-121}
\label{tab:loc_results_xray_densenet_supp}
\begin{adjustbox}{width=\columnwidth,center}
\begin{small}
\begin{sc}
\begin{tabular}{lcccccc}
\toprule
Method & $IoU_{85}$& $IoU_{90}$& $IoU_{95}$ & $IoU_{98}$ & NCC \\
\midrule
\midrule
Gradient&0.173&0.159&0.127&0.081&0.267\\
IG&0.136&0.123&0.095&0.074&0.181\\
GradCAM&0.232&0.223&0.174&0.085&0.344\\
Mgen&0.276&0.264&0.202&0.105&0.338\\
SAGen&0.274&0.255&0.191&0.107&0.337 \\
\midrule
SSyE w/o St.&0.213&0.191&0.139&0.070&0.284\\
SSyE&0.242&0.223&0.168&0.093& 0.378\\
\midrule
CyCSAE w/o St&0.263&0.267&0.222&0.135&0.414\\
CyCSAE&0.268&0.272&0.228&0.141&0.424 \\
\midrule
CyCE &0.265&0.251&0.205&0.111&0.393\\
\midrule
SyCE w/o St.&0.277&0.271&0.221&0.130&0.428\\
SyCE &\textbf{0.290}&\textbf{0.284}&\textbf{0.235}&\textbf{0.144}&\textbf{0.460} \\
\bottomrule
\end{tabular}
\end{sc}
\end{small}
\end{adjustbox}
\end{subtable}
\end{table}
We compare the different embodiments proposed with state-of-the-art methods.
 \begin{itemize}
     \item SyCE outperforms all other approaches (as shown in the main paper).
     \item For SSyE, CyCSAE and SyCE, we also show the localization results when the visual explanation is defined as ${\cal E}(x) = |x - g_{f_c}^{a\star}(x)|$ without the stable generation. These case are denoted with "\textsc{w/o St.}". In all embodiments concerned, the stable image slightly improves the localization results.
     \item In Table \ref{tab:loc_results_xray_resnet_supp}, we also emphasize that the cycle consistency terms $L_d^{cy}$ and $L_{f_c}^{cy}$ improve the localization results of SyCE (compared to "\textsc{w/o} $L^{cy}$").
     \item We note that SSyE produces slightly poorer localization results but remains competitive with state-of-the-art methods.
     \item CyCSAE generates visual explanation competitive with state-of-the-art in the brain MRI problem or even comparable to SyCE in the chest X-Rays task.
 \end{itemize} 
%  For SSyE, CyCSAE and SyCE, we also show the localization results when the visual explanation is defined as ${\cal E}(x) = |x - g_{f_c}^{a\star}(x)|$ without the stable generation. These case are denoted with "\textsc{w/o St.}". In all embodiments concerned, the stable image slightly improves the localization results. In Table \ref{tab:loc_results_xray_resnet}, we also emphasize that the cycle consistency terms $L_d^{cy}$ and $L_{f_c}^{cy}$ improve the localization results of SyCE (compared to "\textsc{w/o} $L^{cy}$"). Then, we can note that SSyE produces localization results competitive with state-of-the-art (or slightly poorer). Finally, CyCSAE generates visual explanation competitive with state-of-the-art in the brain MRI problem or even comparable to SyCE in the chest X-Rays task.
% \newpage
\begin{table}[h!]
\caption{\textbf{Localization results for Brain MRI tumor localization}. Comparison with state-of-the-art techniques and other embodiments (SSyE and CyCSAE).}
\begin{subtable}[h]{0.48\textwidth}
\caption{Results for ResNet-50}
\label{tab:loc_results_mri_resnet_supp}
\begin{adjustbox}{width=\columnwidth,center}
\begin{small}
\begin{sc}
\begin{tabular}{lcccc}
\toprule
Method &$IoU_{97}$& $IoU_{98}$ & $IoU_{99}$ & NCC \\
\midrule
\midrule
Gradient&0.156&0.154&0.131&0.330\\
IG&0.239&0.238&0.196&0.444\\
GradCAM&0.198&0.173&0.115&0.389\\
BBMP&0.270&0.290&0.263&0.409\\
Mgen&0.307&0.319&0.274&0.448\\
SAGen &  0.310&0.330&0.284&0.515 \\
\midrule
SSyE w/o St.&0.255&0.253&0.208&0.441 \\
SSyE &0.275&0.277&0.229&0.462 \\
\midrule
CyCSAE w/o St.&0.258&0.287&0.270&0.463 \\
CyCSAE &0.272&0.305&0.287&0.477 \\
\midrule
CyCE&0.308&0.322&0.270&0.516 \\
\midrule
SyCE w/o St.&0.383&0.406&0.345&0.609\\
SyCE &\textbf{0.389}&\textbf{0.411}&\textbf{0.348}&\textbf{0.615}\\
\bottomrule
\end{tabular}
\end{sc}
\end{small}
\end{adjustbox}
\end{subtable}
\begin{subtable}[h]{0.48\textwidth}
\caption{Results for DenseNet-121}
\label{tab:loc_results_mri_densenet_supp}
\begin{adjustbox}{width=\columnwidth,center}
\begin{small}
\begin{sc}
\begin{tabular}{lcccc}
\toprule
Method& $IoU_{97}$& $IoU_{98}$ & $IoU_{99}$ & NCC \\
\midrule
\midrule
Gradient&0.136&0.128&0.099&0.261\\
IG& 0.210&0.206&0.168&0.397\\
GradCAM&0.282&0.220&0.111&0.342\\
Mgen& 0.333&0.333&0.284&0.519\\
SAGen&0.262&0.264&0.222& 0.440  \\
\midrule
CyCSAE w/o St.&0.222&0.257&0.258&0.404 \\
CyCSAE &0.222&0.258&0.262&0.406 \\
\midrule
CyCE &0.245&0.264&0.236&0.424\\
\midrule
SyCE w/o St.&0.324&0.350&0.310&0.558\\
SyCE &\textbf{0.345}&\textbf{0.372}&\textbf{0.329}&\textbf{ 0.582}  \\
\bottomrule
\end{tabular}
\end{sc}
\end{small}
\end{adjustbox}
\end{subtable}
\end{table}

\newpage\newpage
\subsection{Computation time}
At test time, we generate visual explanations for the different methods either on a NVIDIA MX130 GPU or on 8 cpus Intel 52 Go RAM 1 V100 (as for the optimization). Table \ref{tab:comutation-time_supp} shows the computation time for the different techniques on V100. 
\begin{table}[h!]
\caption{\textbf{Computation time}. Average time (in seconds) of visual explanation generation for LeNet on MNIST digits and ResNet-50 on both X-Ray Pneumonia detecion and Brain MRI tumor localization.}
\label{tab:comutation-time_supp}
\begin{center}
\begin{small}
\begin{sc}
\begin{tabular}{lccc}
\toprule
Method & Digits & Pneumonia & Tumor Loc. \\
\midrule
\midrule
% Gradient&&2.26 $\pm$0.66&2.05 $\pm$ 0.55 \\
% IG& &4.06 $\pm$0.856 & 3.90 $\pm$ 0.541\\
% GradCAM&&0.50 $\pm$ 0.11 & 0.55 $\pm$ 0.10\\
% BBMP& & 17.26 $\pm$ 1.19 & 15.38 $\pm$ 1.12\\
% Mgen& &0.06 $\pm$ 0.09&0.11 $\pm$0.09\\
% SAGen&0.03 $\pm$ 0.003&0.04 $\pm$ 0.004&0.04 $\pm$ 0.01 \\
% \midrule
% CyCE&0.031 $\pm$ 0.08&0.11$\pm$0.07&0.11 $\pm$ 0.09\\
% \midrule
% SyCE w/o St.&0.03 $\pm$ 0.08&0.11 $\pm$0.07&0.11 $\pm$ 0.08\\
% SyCE&0.05 $\pm$0.08&0.14 $\pm$0.07& 0.14 $\pm$ 0.09\\
Gradient&0.09&2.26 &2.05 \\
IG& 0.06&4.06  & 3.90\\
GradCAM&0.02&0.50  & 0.55\\
BBMP& 1.26& 17.26  & 15.38 \\
Mgen& 0.03&0.06 &0.11\\
SAGen&0.03 &0.04 &0.04 \\
\midrule
CyCE&0.03 &0.11&0.11 \\
\midrule
SyCE w/o St.&0.03&0.11 &0.11 \\
SyCE&0.05 &0.14 & 0.14 \\
\bottomrule
\end{tabular}
\end{sc}
\end{small}
\end{center}
\end{table}

% \end{document}
% \documentclass[../main.tex]{subfiles}
% \graphicspath{{\subfix{images/}}}

% %----------------------------------------------------------------------------------------
% %	Illustrations
% %----------------------------------------------------------------------------------------

% \begin{document}
\section{More illustrations}\label{subsec:addillustrations}
Additional illustrations are provided in the following.

\begin{figure*}[ht]
\vskip 0.05in
\begin{center}
\centerline{\includegraphics[width=\textwidth]{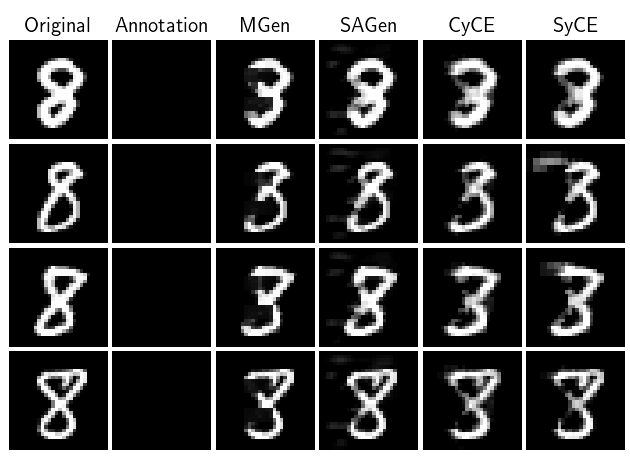}}
\caption{Digits Identification Problem - LeNet - Generation $\chi_1 \longrightarrow \chi_0$ ($8 \longrightarrow 3$) - Comparison of the generated adversarial images with other methods.}
\label{fig:dt_digits_supp}
\end{center}
\vskip -0.2in
\end{figure*}

\begin{figure*}[ht]
\vskip 0.05in
\begin{center}
\centerline{\includegraphics[width=\textwidth]{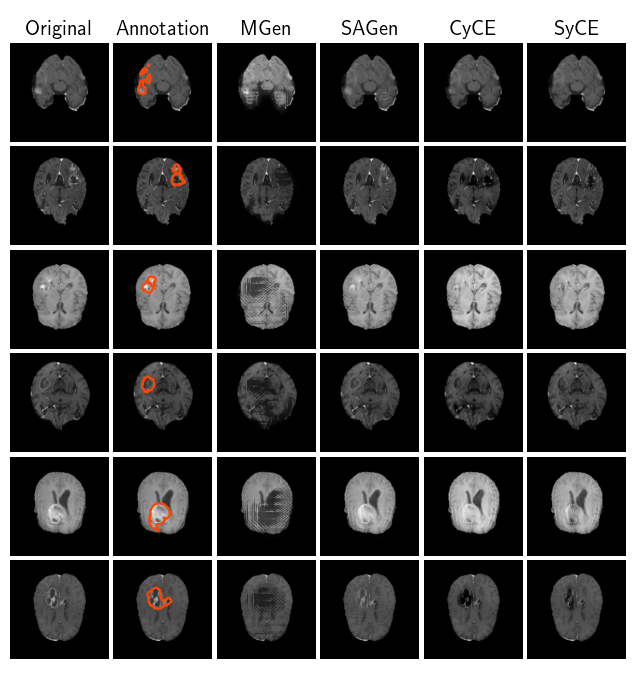}}
\caption{Brain Tumor Localization Problem - ResNet 50 - Generation $\chi_1 \longrightarrow \chi_0$ (Tumor $ \longrightarrow $ No tumor) - Comparison of the generated adversarial images with other methods.}
\label{fig:dt_brain_supp}
\end{center}
\vskip -0.2in
\end{figure*}

\begin{figure*}[ht]
\vskip 0.05in
\begin{center}
% \centerline{\includegraphics[width=\textwidth]{dt_xr_new.png}}
\centerline{\includegraphics[width=\textwidth]{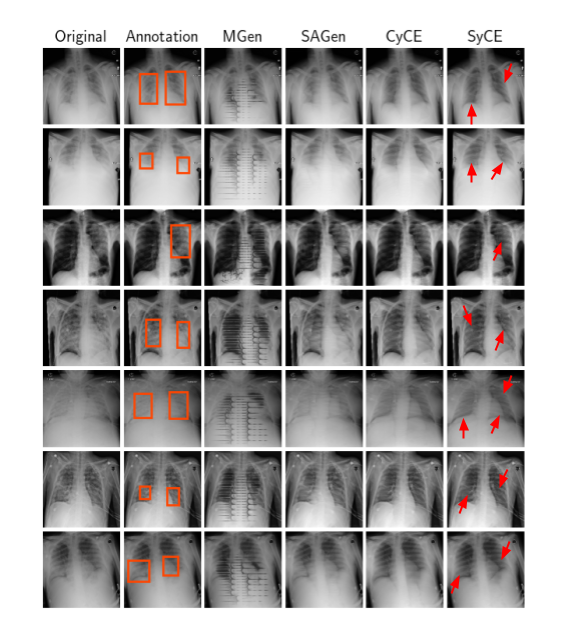}}
\caption{Chest X-rays Problem - ResNet 50 - Generation $\chi_1 \longrightarrow \chi_0$ (Pathological $ \longrightarrow $ Healthy ) - Comparison of the generated adversarial images with other methods.}
\label{fig:dt_xr_supp}
\end{center}
\vskip -0.2in
\end{figure*}

\textbf{Domain translation $\chi_1 \longrightarrow \chi_0$- }Figures \ref{fig:dt_digits_supp}, \ref{fig:dt_brain_supp} and \ref{fig:dt_xr_supp} show some examples of adversarial generation for the three problems in direction: 8 $\longrightarrow$ 3 for digits; Pathological $\longrightarrow$ Healthy for X-Rays and MRI problems. 

\begin{figure*}[ht]
\vskip 0.05in
\begin{center}
\centerline{\includegraphics[width=\textwidth]{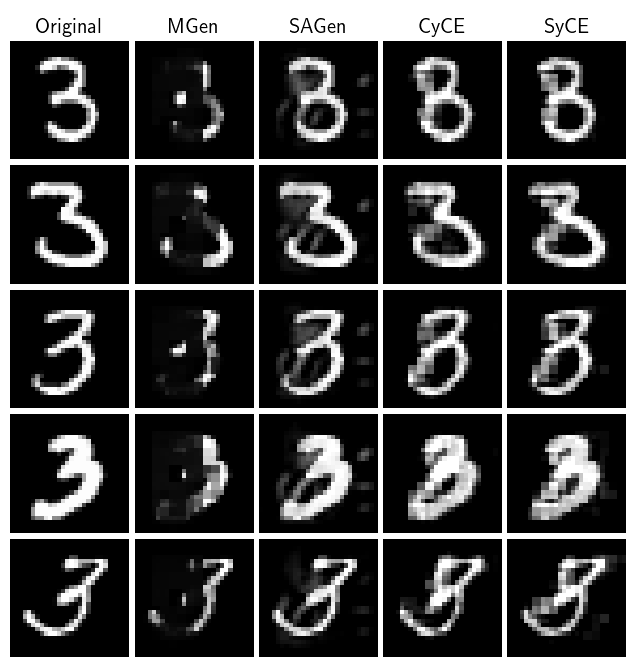}}
\caption{Digits Identification Problem - LeNet - Generation $\chi_0 \longrightarrow \chi_1$ ($3 \longrightarrow 8$) - Comparison of the generated adversarial images with other methods.}
\label{fig:dt_digits_01_supp}
\end{center}
\vskip -0.2in
\end{figure*}

\begin{figure*}[ht]
\vskip 0.05in
\begin{center}
\centerline{\includegraphics[width=\textwidth]{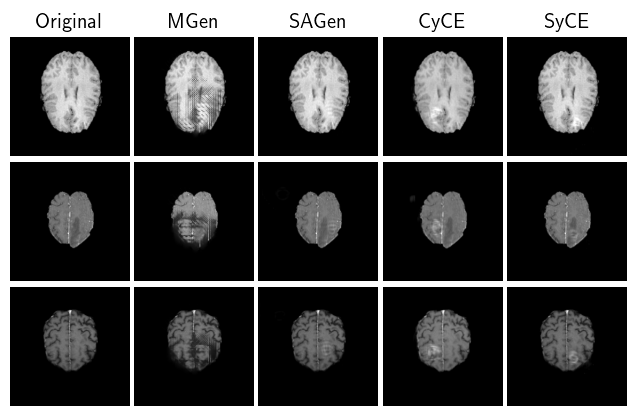}}
\caption{Brain Tumor Localization Problem - ResNet 50 - Generation $\chi_0 \longrightarrow \chi_1$ (No tumor $ \longrightarrow $ Tumor) - Comparison of the generated adversarial images with other methods.}
\label{fig:dt_brain_01_supp}
\end{center}
\vskip -0.2in
\end{figure*}

\begin{figure*}[ht]
\vskip 0.05in
\begin{center}
\centerline{\includegraphics[width=\textwidth]{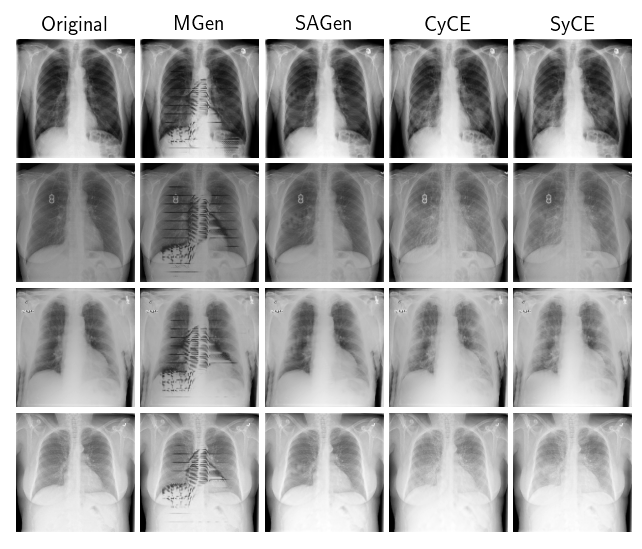}}
\caption{Chest X-rays Problem - ResNet 50 - Generation $\chi_0 \longrightarrow \chi_1$ (Healthy $ \longrightarrow $ Pathological ) - Comparison of the generated adversarial images with other methods.}
\label{fig:dt_xr_01_supp}
\end{center}
\vskip -0.2in
\end{figure*}

\textbf{Domain translation $\chi_0 \longrightarrow \chi_1$- }Figures \ref{fig:dt_digits_01_supp}, \ref{fig:dt_brain_01_supp} and \ref{fig:dt_xr_01_supp} present some examples of adversarial generation for the three problems in direction  3 $\longrightarrow$ 8 for digits; Healthy $\longrightarrow$ Pathological for X-Rays and MRI problems. As mentioned in the paper, we observe on Brain MRI that the same (or very similar) region is perturbated to create a pathology (see Figure \ref{fig:dt_brain_01_supp}). The generator $g_0$ does not express the variability of the pathology. This is less obvious in the other two problems (see Figure \ref{fig:dt_digits_01_supp} and \ref{fig:dt_xr_01_supp}). Note that all generation / perturbation techniques have the same problem.

\begin{figure*}[ht]
\vskip 0.05in
\begin{center}
\centerline{\includegraphics[width=0.8\textwidth]{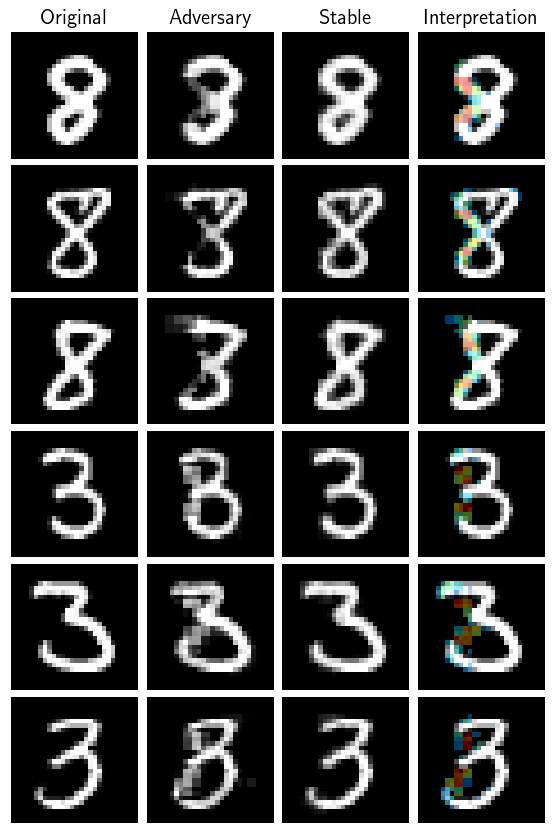}}
\caption{\textbf{Adversary, stable and visual explanation generations}. From left to right for MNIST digit classification: the original image; the adversarial image (for the classifier to explain); the stable image; our resulting explanation map of the classifier's decision.}
\label{fig:sa_hm_digits_supp}
\end{center}
\vskip -0.2in
\end{figure*}

\begin{figure*}[ht]
\vskip 0.05in
\begin{center}
\centerline{\includegraphics[width=\textwidth]{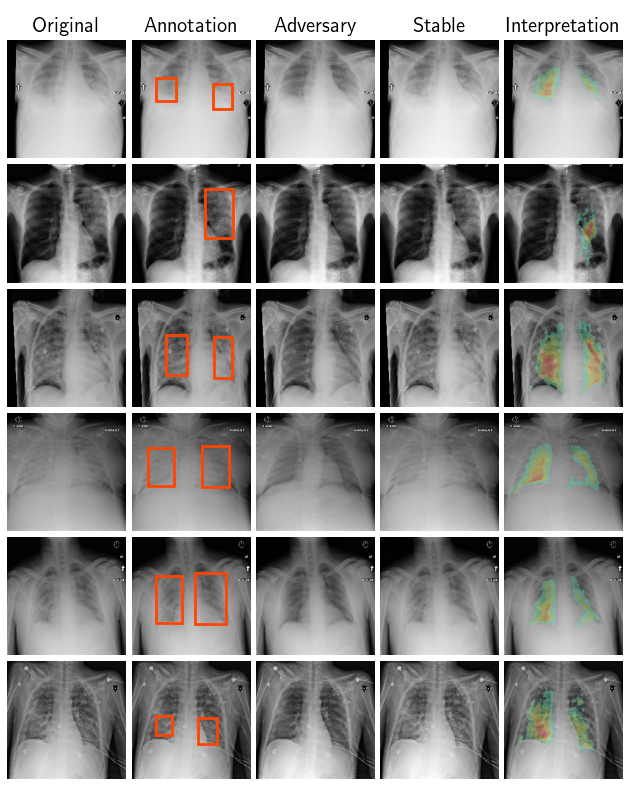}}
\caption{\textbf{Adversary, stable and visual explanation generations}. From left to right for Chest X-Rays pneumonia detection: the original image; the ground annotations pointing out the relevant regions for humans; the adversarial image (for the classifier to explain); the stable image; our resulting explanation map of the classifier's decision.}
\label{fig:sa_hm_xray_supp}
\end{center}
\vskip -0.2in
\end{figure*}

\begin{figure*}[ht]
\vskip 0.05in
\begin{center}
\centerline{\includegraphics[width=\textwidth]{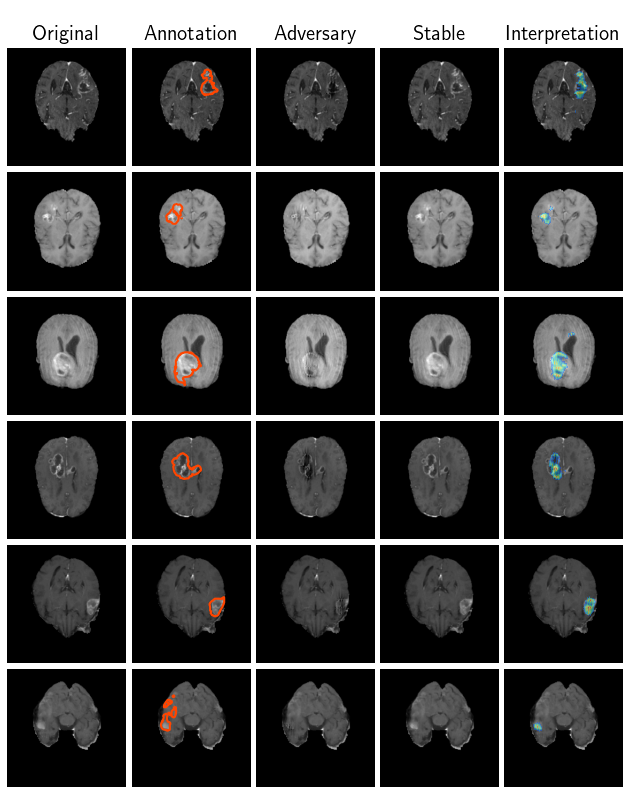}}
\caption{\textbf{Adversary, stable and visual explanation generations}. From left to right for Brain MRI tumor localization: the original image; the ground annotations pointing out the relevant regions for humans; the adversarial image (for the classifier to explain); the stable image; our resulting explanation map of the classifier's decision.}
\label{fig:sa_hm_mri_supp}
\end{center}
\vskip -0.2in
\end{figure*}

\textbf{SyCE generations - }Figures \ref{fig:sa_hm_digits_supp}, \ref{fig:sa_hm_xray_supp} and \ref{fig:sa_hm_mri_supp} display some examples of stable and adversarial images as well as visual explanations generated by SyCE for the three problems.

\begin{figure*}[ht]
\vskip 0.05in
\begin{center}
\centerline{\includegraphics[width=\textwidth]{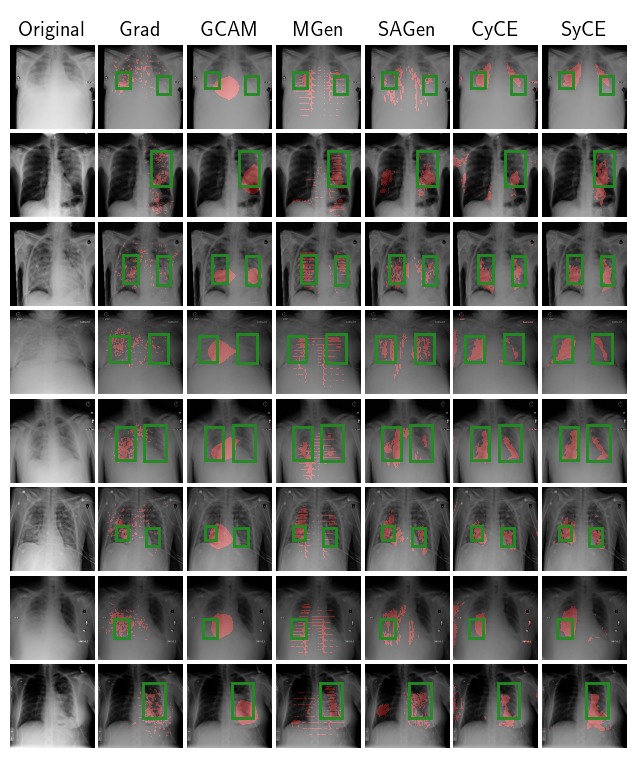}}
\caption{Chest X-rays Problem - ResNet 50 - Comparison with other features attribution methods and against ground truth annotation. For the visualization, all the explanation maps are thresholded at the 95th percentile (representative of the size of the ground truth annotation over the training database)}
\label{fig:hm_xr_supp}
\end{center}
\vskip -0.2in
\end{figure*}

\begin{figure*}[ht]
\vskip 0.05in
\begin{center}
\centerline{\includegraphics[width=\textwidth]{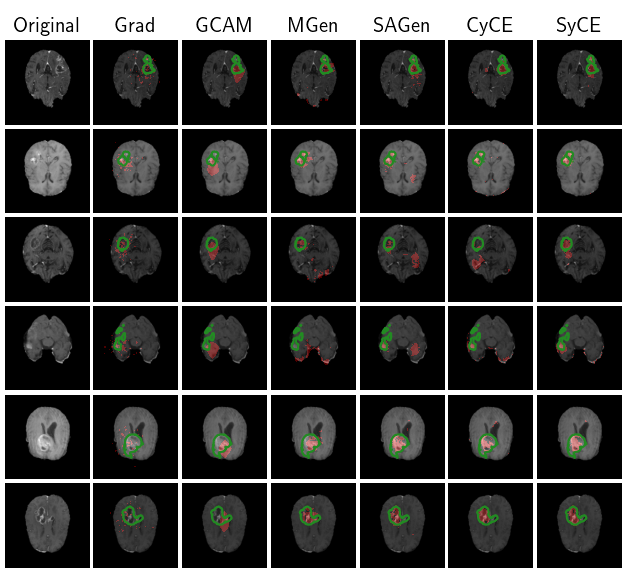}}
\caption{Brain Tumor Localization Problem - ResNet 50 - Comparison with other features attribution methods and against ground truth annotation. For the visualization, all the explanation maps are thresholded at the 98th percentile (representative of the size of the ground truth annotation over the training database)}
\label{fig:hm_brain_supp}
\end{center}
\vskip -0.2in
\end{figure*}

\textbf{Relevant region localization- }Figures \ref{fig:hm_xr_supp} and \ref{fig:hm_brain_supp} compare visual explanation generated by CyCE and SyCE against state-of-the-art methods. Explanation maps are thresholded and compared to human expert annotations.

% \end{document}

% \bibliography{bibliography}

% \end{document}

\end{document}